\documentclass[journal]{IEEEtran}
\ifCLASSINFOpdf
\else
\fi

\usepackage{graphicx}
\usepackage{amsmath}
\usepackage{amssymb}

\usepackage{caption}
\usepackage{makecell}
\usepackage{booktabs}
\usepackage{subfig}
\usepackage{multirow}

\usepackage{multicol}
\usepackage{enumitem}

\usepackage{url}

\usepackage{breakurl}
\usepackage[breaklinks]{hyperref}

\usepackage{algorithmic}
\usepackage{algorithm}

\usepackage{adjustbox} % align graphics

\usepackage{xcolor}
\usepackage{tabu}

\begin{document}
%\title{MAAD-Face: A Massive Attribute Annotation Dataset for Face Images}
\title{MAAD-Face: A Massively Annotated Attribute Dataset for Face Images}

\author{Philipp~Terh\"orst,
        Daniel~F\"ahrmann,
        Jan~Niklas~Kolf,
        Naser~Damer,
        Florian~Kirchbuchner,
        and~Arjan~Kuijper% <-this % stops a space
\thanks{All authors are with the Fraunhofer Institute for Computer Graphics Research IGD, Darmstadt, Germany and with the Technical University of Darmstadt, Darmstadt, Germany. e-mail: \{forename.lastname@igd.fraunhofer.de\}}}% <-this % stops a space
%\thanks{e-mail: \{forename.lastname@igd.fraunhofer.de\}}% <-this % stops a space


\markboth{Journal of \LaTeX\ Class Files,~Vol.~14, No.~8, August~2015}%
{Shell \MakeLowercase{\textit{et al.}}: Bare Demo of IEEEtran.cls for IEEE Journals}



\maketitle

\begin{abstract}
Soft-biometrics play an important role in face biometrics and related fields since these might lead to biased performances, threaten the user's privacy, or are valuable for commercial aspects.
Current face databases are specifically constructed for the development of face recognition applications.
Consequently, these databases contain a large number of face images but lack in the number of attribute annotations and the overall annotation correctness.
In this work, we propose a novel annotation-transfer pipeline that allows to accurately transfer attribute annotations from multiple source datasets to a target dataset.
The transfer is based on a massive attribute classifier that can accurately state its prediction confidence.
Using these prediction confidences, a high correctness of the transferred annotations is ensured.
Applying this pipeline to the VGGFace2 database, we propose the MAAD-Face annotation database.
It consists of 3.3M faces of over 9k individuals and provides 123.9M attribute annotations of 47 different binary attributes.
Consequently, it provides 15 and 137 times more attribute annotations than CelebA and LFW.
%In this work, we propose MAAD-Face, a new face annotations database that is characterized by the large number of its high-quality attribute annotations.
%\textcolor{blue}{MAAD-Face is built on the VGGFace2 database and thus, consists of 3.3M faces of over 9k individuals.}
%\textcolor{blue}{Using a novel annotation-transfer pipeline that allows to accurately transfer from multiple source dataset attributes to a target dataset, MAAD-Face consists of 123.9M attribute annotations of 47 different binary attributes.}
%Consequently, it provides 15 and 137 times more attribute annotations than CelebA and LFW.
%\textcolor{blue}{The proposed annotation-transfer pipeline takes the attribute annotations of the source databases and transfers these to the images of the target database.
%The transfer is based on a massive attribute classifier that can accurately state its prediction confidence.
%Using this prediction confidences a high correctness of the transferred annotations is ensured.
%}
Our investigation on the annotation quality by three human evaluators demonstrated the superiority of the MAAD-Face annotations over existing databases.
Additionally, we make use of the large number of high-quality annotations from MAAD-Face to study the viability of soft-biometrics for recognition, providing insights into which attributes support genuine and imposter decisions.
The MAAD-Face annotations dataset is publicly available.
\end{abstract}


%The proposed annotation-transfer pipeline takes the attribute annotations of the source databases and transfers these to the images of the target database.
%The transfer is based on a massive attribute classifier that can accurately state its prediction confidence.
%Using this prediction confidences a high correctness of the transferred annotations is ensured.

% \fi
\IEEEpeerreviewmaketitle

\section{Introduction}

Soft-biometric characteristics play a major role in face recognition research and applications \cite{DBLP:journals/tifs/DantchevaER16}.
Recently, there is a high interest in studying these attributes and mitigating their effects on recognition performances for fair face recognition systems \cite{9086771}.
Soft-biometrics are also a key factor for privacy-enhancing face recognition technologies, either by recognizing individuals based on soft-biometrics only \cite{7139096} or by suppressing privacy-sensitive characteristics to prevent function creep \cite{DBLP:journals/access/TerhorstRDRBKSK20}.
However, most of these research efforts focus on demographic aspects only.
One possible reason can be the lack of annotated data.
Recent face databases are specifically constructed for the development of face recognition systems.
Consequently, these contain large numbers of faces under diverse settings but lack annotations.

This work closes this gap by proposing the MAAD-Face dataset.
MAAD-Face is a novel face annotations database that is characterized by its large number of high-quality face annotations.
Utilizing our novel annotation-transfer pipeline, we transfer the attribute annotations from two source databases (LFW \cite{LFWTech} and CelebA \cite{DBLP:conf/iccv/LiuLWT15}) to the target database VGGFace2 \cite{Cao18}.
The pipeline trains a massive attribute classifier (MAC) per source database to accurately predict the attributes of the source.
Since the MAC makes use of prediction reliabilities \cite{BTAS_terhoerst}, the pipeline neglects annotations origin from less-confident predictions and thus, ensures a high correctness of the transferred annotations.
MAAD-Face consists of 3.3M faces of over 9k individuals, which is significantly higher than related annotated datasets such as CelebA (0.2M faces of 10k individuals with 40 different attributes) and LFW (13.2k faces of 5.7k individuals with 74 different attributes).
With 123.9M attribute annotations of 47 different binary attributes, MAAD-Face provides 15 and 137 times more attribute annotations than CelebA and LFW.
To analyse the quality of the attribute annotations, three human evaluators investigated the correctness of the annotations of CelebA, LFW, and MAAD-Face.
The results demonstrate the superiority of the MAAD-Face annotations over the other databases.
Finally, we investigated the viability of using soft-biometrics attributes for recognition using MAAD-Face.
We show the relevance of each attribute for genuine and imposter decisions and analyse how many of the most important attributes are necessary to achieve a certain recognition performance.
The MAAD-Face dataset is publicly available under the following link\footnote{\url{https://github.com/pterhoer/MAAD-Face}}.

%Since the MAAD-Face attribute annotations are derived from face embeddings, ...
%four contributions:
%(1) new database with 38.3M attribute annotations
%(2) human evaluation of the annotation quality of MAAD-Face and similar databases such as LFW and CelebA --> demonstrate that the MAAD-Face provide annotations of much higher quality
%(3) an analysis of which and how well such soft-biometric characteristics can be used for verification and identification
%(4) novel annotation-transfer pipeline
%
%
%contributions as a list
%
%
%Soft-biometrics attributes enable recognition systems that preserve the user's privacy...
%previous datasets are designed for face recognition only,
%however recent work has shown that soft-biometric attribute might affect the face recognition performance
%
%abstract
%
%
%
%The results indicate that soft-biometric attributes are a valuable 

To summarize, this work presents \textit{four main contributions}:
\begin{enumerate}
\item A novel annotation-transfer pipeline is proposed that can transfer attribute annotations from multiple source databases to a target database while ensuring a high correctness of the transferred annotations.
We use this pipeline to create MAAD-Face.
\item We propose the MAAD-Face annotations dataset based on VGGFace2 \cite{Cao18}.
MAAD-Face is a new face annotations database consisting of 123.9M attribute annotations of 47 different binary attributes. 
It provides 15 and 137 times more annotations than CelebA and LFW, while the attribute annotations are of higher quality.
\item The third contribution is a human evaluation of the annotation correctness of three large-scale annotation face databases, LFW, CelebA, and MAAD-Face.
These demonstrate the superiority of the MAAD-Face annotations over the other investigated databases.
\item The last contribution is a study on how well these facial attributes can be used for  verification and identification based on soft-biometrics only.
\end{enumerate}

The rest of the paper is structured as follows.
Section \ref{sec:RelatedWork} provides an overview of annotated face datasets and a human evaluation of the annotation-correctness of three highly-annotated datasets.
In Section \ref{sec:annotationTransferPipeline}, the annotation-transfer pipeline is explained and how this is used to create MAAD-Face.
Section \ref{sec:MAADFaceStatistics} provides statistical properties of MAAD-Face and in Section \ref{sec:SoftBiometricRecognition}, the soft-biometric annotations of MAAD-Face are used to evaluate how well these attributes can be utilized to recognize individuals.

\section{Related Works}
\label{sec:RelatedWork}

\subsection{Review of Annotated Face Datasets}
\label{sec:ReviewFaceDatasets}

\begin{table*}[]
\renewcommand{\arraystretch}{1.2}
\centering
\caption{Statistics of related face annotation databases. Distinctive attributes refers to the number of different attributes that are annotated while the number of annotations refers to the total number of (attribute) annotations in the database. To make the number of attributes between different databases comparable, categorical attributes are transformed to binary via one-hot encoding. These are marked a (*). Compared to related databases, MAAD-Face provides the highest number of attribute annotations.} 
\label{tab:DatabaseComparison}
\begin{tabular}{lrrrr}
\Xhline{2\arrayrulewidth}
                &                    &                  & \multicolumn{2}{c}{Attribute annotations}                                                        \\
                \cmidrule(rl){4-5}
Database        & Num. of subjects & Num. of images & Distinctive attributes & Total number of annotations  \\
\hline
ColorFeret \cite{ColorFERET}      & 1.2k               & 14.1k            & 13*                    &       0.2M               \\
Adience \cite{Eidinger:2014:AGE:2771306.2772049}        & 2.3k               & 26.6k            & 10*                    &            0.3M            \\
Morph \cite{1613043} & 13.6k & 55.1k & 10* & 0.6M  \\
VGGFace \cite{DBLP:conf/bmvc/ParkhiVZ15}         & 2.6k               & 2.6M             & 1                      & 2.6M                        \\
VGGFace2 \cite{Cao18}       & 9.1k               & 3.3M            & 11                     & 3.6M                     \\
LFW \cite{LFWTech}             & 5.7k               & 13.2k              & \textbf{74}             & 0.9M               \\
CelebA \cite{DBLP:conf/iccv/LiuLWT15}         & 10.0k                & 0.2M             & 40                     & 8.0M                 \\
MAAD-Face (this paper) & 9.1k               & 3.3M             & 47                     & \textbf{123.9M}           \\
\Xhline{2\arrayrulewidth}  
\end{tabular}
\end{table*}

In recent years, many face databases have been released.
These mainly aimed at providing a large dataset for developing face recognition solutions.
With the use of deep-learning techniques in face recognition, the required data for training these solutions has grown strongly and thus, the sizes of face databases.
However, less attention was given to the estimation of facial attributes.
These soft-biometric characteristics can be of high importance in applications such as access control \cite{DBLP:journals/tifs/DantchevaER16}, human-computer interaction \cite{BTAS_terhoerst}, and law enforcement \cite{DBLP:journals/pami/GengZS07}.
Current face databases only provide insufficient numbers of training annotations for training accurate solutions.
Moreover, these annotations often lack in their correctness and thus, prevent the development of soft-biometric solutions.
In the following, we discuss popular face databases that also contain attribute information.

ColorFeret \cite{ColorFERET} consists of 14.1k images of 1.2k different individuals with different poses under controlled conditions.
The dataset includes a variety of face poses, facial expressions, and lighting conditions.
Each image contains annotations of the individual's gender, ethnicity, head pose, age, glasses, and beard.
In total, ColorFeret provides around 183k soft-biometric annotations.

The Adience dataset \cite{Eidinger:2014:AGE:2771306.2772049} consists of over 26.5k images of over 2.2k different individuals in unconstrained environments.
In total, the dataset provides around 263k annotations for gender and age.
These images were manually annotated.

The Morph dataset \cite{1613043} contains 55.1k frontal face images of more than 13.6k individuals.
For each image, it provides information about the person's gender, ethnicity, age, beard, and glasses.
80.4\% of the faces belong to the ethnicity black, 19.2\% to white, and 0.4\% to others.
The individuals' age varies from 16-77 years.
79.4\% of the faces are within an age-range of $[20,50]$.
In total, the Morph database provides over 0.5M annotations for soft-biometric attributes.

VGGFace \cite{DBLP:conf/bmvc/ParkhiVZ15} and VGGFace2 \cite{Cao18} are two databases from the University of Oxford.
VGGFace \cite{DBLP:conf/bmvc/ParkhiVZ15} contains 2.6M images from 2.6k individuals and provides information about the head pose (frontal, profile).
VGGFace2 \cite{Cao18} contains faces from over 9k subjects with over 3M images.
The dataset contains a large variety of pose, age, and ethnicity.
Over 40\% of the face are frontal and over 50\% are half-frontal.
Most images belong to individuals over 18 years old and around 40\% belong to the age group of $[25,34]$.
For each image, gender annotations are available.
A subset of 30k images of celebrities was additionally annotated with 10 further attributes about the individual's hair, beard, glasses, and hat.
In total, VGGFace2 provides 3.6M annotations about the person's face.

Labelled Faces in the Wild (LFW) \cite{LFWTech} contains 13.2k images of 5.7k different identities from unconstrained environments.
It contains variability in pose, lighting, expression, and demographics.
With 74 binary attributes, it provides a large diversity on binary attribute annotations, such as attributes belonging to demographics, hair, skin, accessories, and capture environment.
However, as we will show in Section \ref{sec:EvaluatingannotationCorrectness}, the correctness of these annotations are often weak (72\% accuracy compared to human annotations).
In total, LFW provides over 0.9M attribute annotations.
Moreover, it should be mentioned that LFW and VGGFace2 have some overlapping subjects since both databases contain many images of celebrities \cite{herta}.

The CelebFaces Attributes Dataset (CelebA) \cite{DBLP:conf/iccv/LiuLWT15} contains over 202k images of 10.0k different subjects.
It covers large pose variations and background clutter and provides rich annotations for 40 binary attributes.
In total, CelebA provides over 8M annotations for soft-biometric attributes.
These include attributes belonging to demographics, hair, face geometry, and accessories.

A summary of related face annotation databases are shown in Table \ref{tab:DatabaseComparison}.
There, the number of subjects and face images are shown along with the number of attributes and the total number of annotations.

In this work, we propose the MAAD-Face annotation database.
Using our novel annotation-transfer technique we are able to create highly accurate face annotations building upon VGGFace2.
Consequently, it contains over 3.3M face images from over 9.1k different subjects with a large variety of poses, ages, and ethnicities.
MAAD-Face provides annotations for 47 binary attributes.
In total, it consists of over 123.9M attribute annotations, which is over 15 times higher than the second-largest face annotation dataset.

\subsection{Soft-Biometrics from Faces}
Recently, research on the estimation of soft-biometric attributes from face images has shifted from the use of handcrafted features to the use of deep convolutional neural networks \cite{DBLP:journals/tifs/DantchevaER16}.
These kinds of approaches often surpass human-level performance, e.g. for age \cite{6613022}, gender \cite{NIPS1990_405}, or race estimation \cite{6920084}.
Due to the high performance of automatic approaches, several works demonstrated the benefits of soft-biometrics for recognition.
Recent works demonstrated that soft-biometric information only, extracted from face images, can be successfully used to verify \cite{DBLP:conf/eccv/RuddGB16, DBLP:conf/btas/SamangoueiC16} and identify \cite{DBLP:conf/atsip/GhallebSA16, DBLP:conf/isba/AlmudhahkaNH16, DBLP:journals/pami/ReidNS14} individuals.
This is especially useful when facing low-quality capture, such as images from great distances \cite{DBLP:journals/tifs/Tome-GonzalezFVN14}.
Moreover, soft-biometric information can also support primary biometric modalities such as face recognition \cite{DBLP:journals/tifs/Gonzalez-SosaFV18, DBLP:conf/icpr/Gonzalez-SosaVH18, DBLP:journals/cviu/ZhangBDP15}.

The basis for these successes is the developments of accurate soft-biometric estimators.
The solutions cover a wide range of mechanisms such as domain-adaption \cite{Rothe2018, DBLP:journals/pami/HanJWSC18, Rodrguez2017AgeAG, BTAS_terhoerst, DBLP:conf/fusion/TerhorstHKDKK19}, cascade CNN's \cite{7791154, 8007264, MANSANET201680}, autoencoders \cite{ZAGHBANI2018337}, stacked model \cite{7376610}, and deep regression trees \cite{DBLP:journals/corr/abs-1712-07195}.

Taking into account the current pandemic conditions, Alonso-Fernandez et al. \cite{DBLP:conf/biosig/Alonso-Fernandez20} demonstrated that soft-biometric attributes can still be accurately states from faces wearing masks.
In \cite{terhrst2020identity}, Terhörst et al. showed that 74 out of 113 analysed soft-biometric attributes are encoded in face templates demonstrating a strong need for privacy-enhancing methods.
In \cite{DBLP:journals/corr/abs-2103-01592}, it was shown that soft-biometric attributes, in general, have a strong influence on the face recognition performance demonstrating the need for bias mitigating solutions beyond demographics.

%
%RAAGE used for this pipeline
%BEyond Identity \cite{terhrst2020identity} --> privacy and bias 
%sb bias \cite{DBLP:journals/corr/abs-2103-01592}

Building on the reliability measure developed in \cite{BTAS_terhoerst}, this work proposes a novel annotation transfer pipeline that is able to do the transfer task with a high annotation correctness.
This pipeline is used to create the MAAD-Face annotations dataset.
This dataset provides the required data for the developing and evaluating of privacy-enhancing and bias-mitigating face recognition solutions to mitigate the privacy and bias issues mentioned in \cite{terhrst2020identity, DBLP:journals/corr/abs-2103-01592}.
For developing these solutions, it provides the required training data.
For evaluating privacy-enhancing and bias-mitigating face recognition solutions, it provides a solid basis of data that can serve as the test set.

\section{annotation-Transfer Pipeline}
\label{sec:annotationTransferPipeline}

In this section, we will present one of the main contributions of this work, a novel annotation-transfer pipeline that can create highly reliable and accurate attribute annotations.
We will explain this pipeline based on the example of the MAAD-Face annotations database.
The MAAD-Face database was created by transferring the attribute annotations of CelebA and LFW on the images of VGGFace2.

\begin{figure*}
\centering
\includegraphics[width=0.7\textwidth]{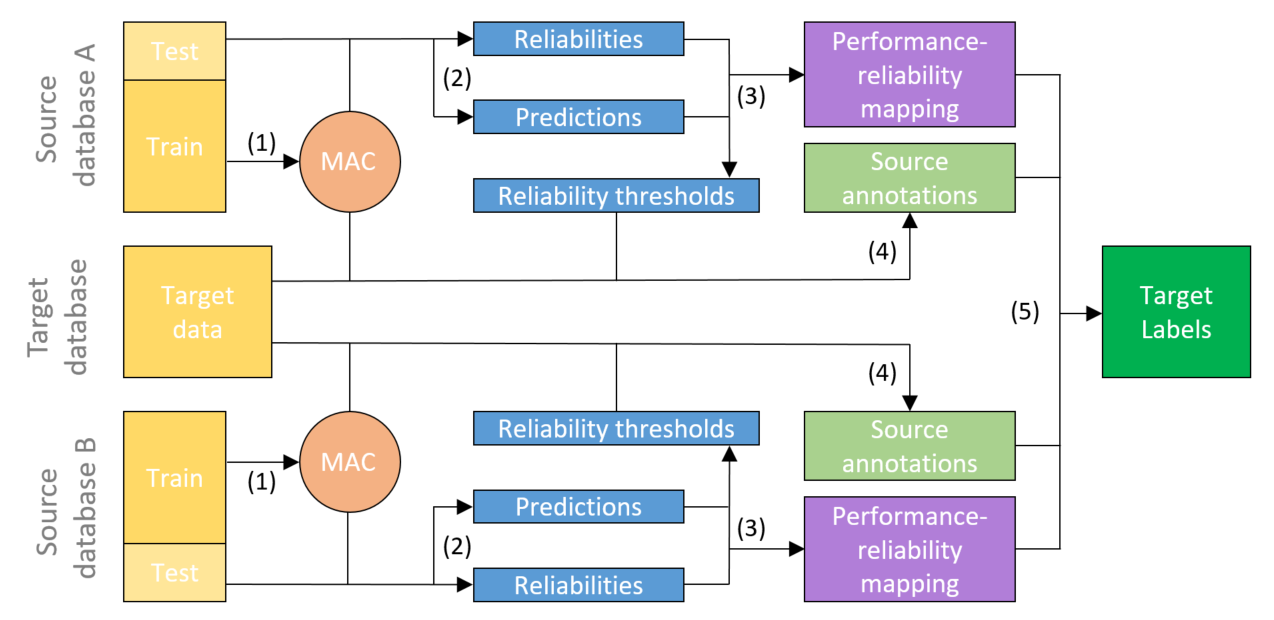}
\caption{Overview of the proposed annotation-transfer pipeline. (1) A MAC is trained on the training part of each source dataset. (2) The MAC produces predictions and prediction reliabilities on the test set. (3) These are used to determine the reliability thresholds per attribute and to calculate the performance-reliability mapping. (4) The MAC and the reliability thresholds are used to create (source) attribute annotations for the target dataset. Finally, (5) the source annotations from each source dataset are aggregated using the corresponding performance-reliability mappings to construct the final target annotations for the target dataset.}
\label{fig:annotation-transfer-pipeline}
\end{figure*}

An overview of the proposed annotation-transfer pipeline is shown in Figure \ref{fig:annotation-transfer-pipeline}.
The pipeline consists of five steps that aim to transfer the annotations of source databases to the target database.
\begin{enumerate}
\item A massive attribute classifier (MAC) is trained on the training part of each source dataset.
Besides making predictions about the estimated annotations of a given image, the MAC is able to additionally providing a reliability statement that states the model's prediction confidence for each annotation.
\item The MAC predicts the annotations on the test-parts of the source datasets including the prediction reliabilities.
\item Based on this performance, the reliability threshold for each attribute is determined. Moreover, a performance-reliability mapping is calculated that allows assigning an attribute-reliability with its expected correctness (performance).
\item The MAC predicts the attribute annotations as well as the corresponding reliabilities for each image in the target dataset.
Predicted annotations below the attribute threshold will be rejected to guarantee a high quality of the transferred source annotations.
\item Finally, the source annotations (with their reliabilities) are aggregated using the corresponding performance-reliability mapping. 
If the source annotations for an image produces different annotations, the annotation is used as the target annotation that has the higher expected correctness.
\end{enumerate}

%First, a massive attribute classifier (MAC) is trained on the training part of the source dataset.
%Besides making predictions about the estimated annotations of a given image, the MAC is able to additionally providing a reliability statement that states the model's prediction confidence for each annotation.
%Second, the MAC predicts the annotations on the test-part of the source dataset including the prediction reliabilities.
%Based on this performance, the reliability threshold for each attribute is determined.
%In the last step, the MAC predicts the attribute annotations as well as the corresponding reliabilities for each image in the target dataset.
%Predicted annotations below the attribute threshold will be rejected to guarantee a high-quality of the transferred annotations.

%The goal of this work was to generate a face database with a high number of highly reliable attributes.
%To achieve this, we transferred annotations from two other strongly-annotated face databases to the large-scale VGGFace 2 dataset and reject transferred annotations of low quality.
%We trained a massive attribute classifier (MAC) on the attributes of LFW and CelebA.
%Then, the MAC made a prediction on each image of the VGGFace 2 dataset.
%For each prediction the MAC additionally provides a reliability statement that states the model's prediction confidence.
%By rejecting annotation predictions with a low reliabilities, we ensure that the remaining annotations are of high quality.

In the following sections, we describe how (a) the MAC training procedure is conducted on the source datasets, (b) the prediction reliability statements of the MAC are calculated, and (c) how this results in the final annotations for the target database.

\subsection{The Massive Attribute Classifier (MAC)}
\label{sec:PreparingMAC}

To transfer the annotations for each attribute from source databases to a target database, we (a) train a MAC jointly on all attributes of a source database to make use of a shared embedding space and (b) construct the MAC such that it can produce accurate reliability measures for each attribute-annotation prediction.

The MAC is a neural network that is trained to predict the attributes of the source dataset.
The network architecture is chosen to maximize the prediction accuracy.
As it will be demonstrated in Section \ref{sec:Reliability}, the only requirement for the MAC is to train with at least one dropout-layer \cite{Srivastava:2014:DSW:2627435.2670313}.
We will need this layer to determine the reliability of a prediction.
Each source database is subject-exclusively divided into a 80\% training set and a 20\% test set.
A separate MAC is trained for each source training set.
To construct MAAD-Face, we use VGGFace2 as the target database and CelebA and LFW as source databases for training two MACs.

In the following, we describe the structure and the training details of the MAC, as well as the data cleaning process used.
As we demonstrated in Section \ref{sec:ReviewFaceDatasets}, many annotations of LFW are wrongly assigned.
To prevent confusion of the MAC trained on these annotations, we filter out annotations that are wrongly assigned with a high probability.

\subsubsection{MAC training}
Generally, the training of the MAC can vary and should be task and data-dependent.
In order to prepare the MAC for our annotation-transfer pipeline, it needs to be trained with at least one dropout-layer \cite{Srivastava:2014:DSW:2627435.2670313} and consists of a soft-max layer as the output.

For the construction of MAAD-Face, we build the MAC on the templates of face images.
As shown in the work of Terh\"orst et al. \cite{terhrst2020identity}, one can easily and accurately predict many attributes from such templates.
Based on these results, we trained a neural network model that takes FaceNet \cite{DBLP:journals/corr/SchroffKP15} embeddings as an input to jointly predict multiple attributes of the source database.
However, a MAC can also be trained end-to-end or by fine-tuning an existing network.
The utilized network structure follows the one used by Terh\"orst et al. \cite{terhrst2020identity}.
It consists of two initial layers, the input layer of size $n_{in}$ and the second dense layer of size 512.
The size of the utilized face embedding is denoted by $n_{in}$ and for our FaceNet model\footnote{\url{https://github.com/davidsandberg/facenet}} refers to 128 dimensions.
Starting from the second layer, each attribute $a$ has an own branch consisting of two additional dense layers of size 512 and $n_{out}^{(a)}$, where $n_{out}^{(a)}$ refers to the number of attributes per class.
Each layer has a ReLU activation, except for the output-layers.
These have softmax activations.
Moreover, Batch-Normalization \cite{DBLP:conf/icml/IoffeS15} and dropout \cite{Srivastava:2014:DSW:2627435.2670313} ($p_{drop} = 0.5$) is applied to every layer.
The dropout allows achieve a generalized performance and also enables us to derive reliability statements about the predictions as we will describe in Section \ref{sec:Reliability}.
The training of the MAC was done in a multi-task learning fashion by applying a categorical cross-entropy loss for each attribute branch and use an equal weighting between each of these attribute-related losses.
For the training, an Adam optimizer \cite{DBLP:journals/corr/KingmaB14} was used with $e=200$ epochs, an initial learning rate $\alpha = 10^{-3}$, and a learning-rate decay of $\beta = \alpha/e$.
The parameter choices followed \cite{terhrst2020identity}.
The batch size $b$ was chosen according to the amount of available training data, $b=1024$ for CelebA and $b=16$ for LFW.

\subsubsection{Cleaning training attribute annotations}
\label{sec:CleaningTrainingannotations}
For the annotation-transfer pipeline, this step is only necessary if a source database consists of attribute annotations of low quality.
As we demonstrated in Section \ref{sec:ReviewFaceDatasets}, this is the case for LFW.
However, the quality of the input data of a model is important for the quality of its output data as demonstrated by Geiger et al. \cite{DBLP:conf/fat/GeigerYYDQTH20}.
Therefore, in this section we will describe an annotation-cleaning process that was used on the LFW dataset.

While in CelebA the attributes are of binary nature, the annotations in LFW originate from the prediction probabilities of a binary classifier \cite{LFWTech}.
Therefore, these annotations are continuous and measure the degree of the attribute \cite{DBLP:conf/iccv/KumarBBN09, DBLP:journals/pami/KumarBBN11}.
Positive values represent "true" annotations and negative values represent "false" annotations.
A positive annotation for an attribute $a$ of an image means that the face in the image has the attribute $a$.
For instance, a face with a positive annotation for \textit{Beard} represents a face with a beard.
In contrast, a negative annotation for an attribute $a$ of an image means that the face in the image has not the attribute $a$. 
However, using the prediction probabilities of a binary classifier does not necessarily reflect the correctness of the prediction as shown in recent works \cite{DBLP:journals/corr/GuoPSW17,NIPS2015_5658,journals/corr/NguyenYC14}.
Consequently, a wide range of the LFW annotations centred around a value of zero is unreliable.
%, and thus, many of these annotations are wrongly assigned.

To ensure that our MAC learns on meaningful LFW-annotations, we manually removed these centred annotations as described in the work of Terh\"orst et al. \cite{terhrst2020identity}.
Therefore, we assigned an upper and lower score threshold for each attribute.
Images with a score over the upper threshold are assigned as true, images with a score under the lower threshold are assigned as false, images with scores within the range are rejected.
The upper and lower thresholds for one attribute are manually determined by moving potential thresholds away from zero.
At each potential threshold, ten images with the closest attribute scores are investigated.
Here, the original LFW annotations of the images are manually investigated for correctness.
If only eight or fewer attributes are investigated as correct, the potential threshold is further moved away from the starting point and the procedure is repeated.
If a potential threshold returns images with 9 or more correct annotations, it is chosen as the limit.
%The correctness of the annotations is evaluated manually.
Repeating this over all attributes will result in a lower and an upper threshold for each of these attributes.
By binaryzing the scores with these upper and lower thresholds, reduces the number of annotations by 51,7\%.
However, it also ensures an error-minimizing data basis of the MAC.
Thus, it allows us to train the MAC on meaningful and mostly correctly annotated data.

\subsection{Deriving reliability statements} 
\label{sec:Reliability}
To ensure that the target database will only get annotations of high quality, the prediction reliability is additionally estimated for each prediction (target annotation).
Therefore, we follow the methodology described in \cite{BTAS_terhoerst} to enable our MAC to accurately state its own prediction confidence (reliability).
To derive the reliability statement additionally to an attribute prediction, $m=100$ stochastic forward passes are performed.
In each forward pass, a different dropout-pattern is applied, resulting in $m$ different softmax outputs $v_i^{(a)}$ for each attribute $a$.
Given the outputs of the $m$ stochastic forward passes of the predicted class $\hat{c}$ denoted as $x^{(a)}=v_{i,\hat{c}}^{(a)}$, the reliability measure is given as 
\begin{align*}
rel(x^{(a)}) =   \dfrac{1-\alpha}{m} \sum_{i=1}^m x_i^{(a)} - \dfrac{\alpha}{m^2} \sum_{i=1}^m \sum_{j=1}^m |x_i^{(a)} - x_j^{(a)}|,
\end{align*}
with $\alpha=0.5$, following the recommendation in \cite{BTAS_terhoerst}.
The first part of the equation is a measure of centrality and utilizes the probability interpretation of the softmax output.
A higher value can be interpreted as a high probability that the prediction is correct.
The second part of the equation is the measure of dispersion and quantifies the agreement of the stochastic outputs $x$.
In \cite{BTAS_terhoerst}, this was shown to be an accurate reliability measure.
%and \cite{DBLP:conf/fusion/TerhorstHKDKK19}

\subsection{Attribute annotation generation}
\label{sec:AttributeannotationGeneration}

In this section, we combine the MAC models of the source datasets and the reliability measure to create high-quality target annotations.
First, we will describe how to set the reliability thresholds for each attribute and MAC.
Then, we will describe how this can be used to create the annotations on the target dataset.

\subsubsection{Defining reliability thresholds}

For each source database, a MAC model $\mathcal{M}$ was already trained on the training part as described in Section \ref{sec:PreparingMAC}.
Now, the MAC predicts the source annotations on the test-part including the prediction reliabilities.
Moreover, the MAC repeats this step on the target database.
For each attribute $a$ of the source database, the reliability threshold $\textit{thr}_{Source}^{(a)}$ is chosen such that the (balanced) prediction accuracy of $a$ is over $acc_{min}$\% and at least $d_{min}$\% of the target samples are over this threshold.
Consequently, $acc_{min}$ defines the quality of the target annotations while $d_{min}$ define the amount of the annotations in the target database.
If an attribute does not accomplish this requirement, the attribute is discarded.

For the creation of MAAD-Face, we set $acc_{min}=90\%$ and $d_{min}=50\%$ to receive a large number of high-quality annotations.
This results in manually chosen reliability thresholds $\textit{thr}_{CelebA}^{(a)}$ and $\textit{thr}_{LFW}^{(a)}$ for each attribute $a \in \mathcal{A}$.

%In this section, we put the MAC model $\mathcal{M}$ and the reliability measure together to create the MAAD-Face database.
%
%First, the CelebA and (cleaned) LFW datasets are randomly split into a 80\% training and 20\% testing set that are subject-disjoint.
%The model $\mathcal{M}$ is further trained on both training sets to predict their dataset-related attributes as described in Section \ref{sec:PreparingMAC}.
%The test sets are used to manually determine the reliability threshold for each attribute.
%These thresholds are chosen such that the (balanced) prediction accuracy for this attribute is over 90\% and at least 50\% of the test samples are over this threshold.
%If this is not achieved for any reliability threshold, the attribute is discarded.
%This results in manually chosen reliability thresholds $\textit{thr}_{CelebA}^{(a)}$ and $\textit{thr}_{LFW}^{(a)}$ for each attribute $a \in \mathcal{A}$.

\subsubsection{Creating target annotations}

After defining the reliability thresholds for each MAC and attribute $a \in \mathcal{A}$, we can create the target annotations.
Therefore, each MAC computes its predictions $p_{Source}$ and prediction reliabilities $r_{Source}$ on the target dataset.
The prediction \textit{True} is defined as 1, the prediction \textit{False} is defined as -1.
If an attribute-prediction $p_{Source}^(a,i)$ for an image $i$ has a prediction-reliability below the threshold $r_{Source}^{(a,i)}<\textit{thr}^{(a)}_{Source}$, the annotation is set to 0 (\textit{undefined}).
In this case, the MAC is not confident enough about its prediction and rejecting these predictions guarantee high-quality remaining annotations.
For each source dataset, this procedure results in a set of annotations $l_{Source}$ for the target dataset images.
Finally, this set of annotations have to be combined to create the target annotations.
If an attribute just appears in one of the source datasets, the source annotations $l_{Source}$ are directly used for the target dataset.
If an attribute appears in multiple source datasets, we have to decide which annotation to use as the target annotation.
In this case, the reliability $r_{Source}$ is mapped back to the performance of the test set $acc(r_{Source})$ and the annotation assigned with the highest map-back performance is used for the target annotation.
Please note that such a decision can not be made based on the reliability-level only since the range of the reliability values vary between each MAC.
Mapping back the reliability values to the test-set performances allow an aligned comparison of the annotation-quality.

Algorithm \ref{algo:annotationGeneration} summarizes the annotation generation procedure.
The inputs are the predictions $\{p_{Source}\}$, the corresponding reliabilities $\{r_{Source}\}$, the reliability thresholds $\{\textit{thr}_{Source}\}$, as well as a set of all attribute $\mathcal{A}$.
The output of the algorithm is the annotations $l_{Target}$ of the target dataset.
The \textit{transfer} function transforms the predictions $p_{Source}$ into the source annotations $l_{Source}$ based on the prediction reliabilities $r_{Source}$ and the corresponding attribute reliability thresholds $\textit{thr}_{Source}$.
If an attribute appears in multiple source databases, the \textit{highest} function maps back the reliability to the test-set performance $acc(r_{Source}^{(a,i)}$ and returns the annotation $l_{Source}^{(a,i)}$ with the highest map-back performance.

The last step (\textit{obtainPlausability}) performs a plausibility check including required corrections, given the target annotations $l_{\textit{Target}}$, the attribute classes $\mathcal{A}$, and the corresponding attributes.
For each class, at maximum one attribute can be true.
For instance, for the class gender, either the attribute male or female can be true.
A list of the classes with the corresponding attributes is shown in Table \ref{tab:AttributePerformanceMAADFace}.
Due to this restriction, we set all attribute annotations for an image $i$ to undefined (0) if more than one attribute showed true before.
This aims at maintaining high-quality annotations.

 \begin{algorithm}
 \caption{ - Annotation Generation}
 \label{algo:annotationGeneration}
 \begin{algorithmic}[1]
 \renewcommand{\algorithmicrequire}{\textbf{Input:}}
 \renewcommand{\algorithmicensure}{\textbf{Output:}}
 \REQUIRE $\{p_{Source}\}, \{r_{Source}\}, \{\textit{thr}_{Source}\}, \mathcal{A}$
 \ENSURE target dataset annotations $l_{\textit{Target}}$
 \FOR{$a \in \mathcal{A}$}
 	\FOR{\textbf{each} source dataset}
 	\STATE $l_{Source}^{(a)} \leftarrow \textit{transfer}(p_{Source}^{(a)}, r_{Source}^{(a)}, \textit{thr}_{Source}^{(a)})$
 	\ENDFOR
 \ENDFOR
 \STATE $l_{\textit{MAAD}} = \textit{zeros}(|\mathcal{A}|, |\mathcal{I}|)$
 \FOR{$a \in \mathcal{A}$}
   \FOR{$i \in \mathcal{I}$} 
   \STATE $l_{Target}^{(a,i)} \leftarrow \textit{highest}(\{l_{Source}^{(a,i)}\}, \{acc(r_{Source}^{(a,i)})\})$
 
   \ENDFOR
 \ENDFOR 
 \STATE $l_{\textit{Target}} \leftarrow \textit{obtainPlausibility}(l_{\textit{Target}}, \mathcal{A} )$
 \RETURN $l_{\textit{Target}}$
 \end{algorithmic}
 \end{algorithm}

\subsection{Discussion}
The proposed annotation transfer pipeline is related to homogenous discrepancy-based domain adaptation methods \cite{DBLP:journals/iet-ipr/MadadiSNHM20, DBLP:journals/ijon/WangD18}.
Since the feature spaces between the source and target domains are identical and only differ in terms of data distribution, it is similar to homogeneous domain adaptations \cite{DBLP:journals/ijon/WangD18}.
Since the reliability measure of the transferred labels can be interpreted as the distance between the source and the target domain, the proposed pipeline is also similar to discrepancy-based domain adaption methods \cite{DBLP:journals/iet-ipr/MadadiSNHM20}.
In contrast to classical domain adaptation methods that utilize labeled data in the source domains to execute new tasks in a target domain \cite{DBLP:journals/ijon/WangD18}, the proposed approach solves the same task in the target domain.
However, it measures the discrepancy (reliability) in the target domain to prevent false decision (annotations).
This might only lead to decisions when the target and source domain share a specific similarity but also ensures a high correctness of the decisions (annotations).

\begin{table}[]
\setlength{\tabcolsep}{4pt}
\renewcommand{\arraystretch}{1.2}
\centering
\caption{Analysing the MAC prediction performance in comparison to Microsoft Azure under different head poses and lighting conditions. The values reported are the balanced accuracies based on ColorFeret under frontal and (artificial) lighting from the side. Moreover, the performance under different headpose from frontal ($0^\circ$) to profile ($90^\circ$). While the MAC classifier can predict 47 attributes, in this evaluation only shared attributes between ColorFeret and Azure are considered. The rejected rate describes the ratio of images that could not be processed by the algorithm. In general, both approaches perform well. However, for non-frontal images and images with side lighting, Azure rejects most of the images without any predictions.}
\label{tab:MAC_SOTA}
\begin{tabular}{llcccccccc}
\Xhline{2\arrayrulewidth}
                &                 & \multicolumn{4}{c}{MAC (ours)} & \multicolumn{4}{c}{Microsoft Azure} \\ \cmidrule(rl){3-6} \cmidrule(rl){7-10}
                &                 & $0^\circ$     & $45^\circ$     & $90^\circ$ & avg   & $0^\circ$       & $45^\circ$      & $90^\circ$  & avg     \\
                \hline
\multirow{9}{*}{\rotatebox[origin=c]{90}{Frontal lighting}} & Young           & 0.75      & 0.81     & 0.77    & 0.78 & 0.55        & 0.51      & 0.49      & 0.52 \\
                & Middle\_Aged    & 0.54      & 0.53     & 0.51    & 0.53 & 0.73        & 0.72      & 0.66      & 0.70 \\
                & Senior          & 0.87      & 0.86     & 0.84    & 0.85 & 0.84        & 0.82      & 0.83      & 0.83 \\
                & Male            & 0.94      & 0.93     & 0.83    & 0.90 & 0.99        & 0.99      & 1.00      & 0.99 \\
                & Eyeglasses      & 0.97      & 0.99     & 0.96    & 0.97 & 0.99        & 0.99      & 1.00      & 0.99 \\
                & Beard           & 0.89      & 0.88     & 0.91    & 0.90 & 0.86        & 0.85      & 0.84      & 0.85 \\
                & Mustache        & 0.98      & 0.97     & 0.90    & 0.95 & 0.56        & 0.55      & 0.61      & 0.58 \\
                & Average (avg)   & 0.85      & 0.85     & 0.82    & 0.84 & 0.79        & 0.78      & 0.78      & 0.78 \\
                \hline
                & Rejected images & 0.00      & 0.00     & 0.02    & 0.01 & 0.04        & 0.13      & 0.85      & 0.34 \\ \Xhline{2\arrayrulewidth}
\multirow{9}{*}{\rotatebox[origin=c]{90}{Side lighting}} & Young           & 0.75      & 0.80     & 0.73    & 0.76 & 0.56        & 0.52      & 0.92      & 0.67 \\
                & Middle\_Aged    & 0.54      & 0.52     & 0.49    & 0.52 & 0.69        & 0.72      & 0.88      & 0.76 \\
                & Senior          & 0.87      & 0.84     & 0.79    & 0.83 & 0.77        & 0.81      & 0.96      & 0.85 \\
                & Male            & 0.89      & 0.92     & 0.86    & 0.89 & 0.99        & 0.99      & 1.00      & 0.99 \\
                & Eyeglasses      & 0.97      & 0.98     & 0.91    & 0.95 & 0.99        & 0.99      & 1.00      & 0.99 \\
                & Beard           & 0.89      & 0.89     & 0.96    & 0.91 & 0.87        & 0.91      & 1.00      & 0.93 \\
                & Mustache        & 0.98      & 0.95     & 0.67    & 0.87 & 0.58        & 0.60      & 0.48      & 0.55 \\
                & Average (avg)   & 0.84      & 0.84     & 0.77    & 0.82 & 0.78        & 0.79      & 0.89      & 0.82 \\
                \hline
                & Rejected images & 0.00      & 0.03     & 0.32    & 0.12 & 0.13        & 0.50      & 0.99      & 0.54\\
\Xhline{2\arrayrulewidth}
\end{tabular}
\end{table}

\section{MAAD-Face}

\subsection{MAAD-Face Statistics}
\label{sec:MAADFaceStatistics}

\begin{figure*}
\centering
\includegraphics[width=0.9\textwidth]{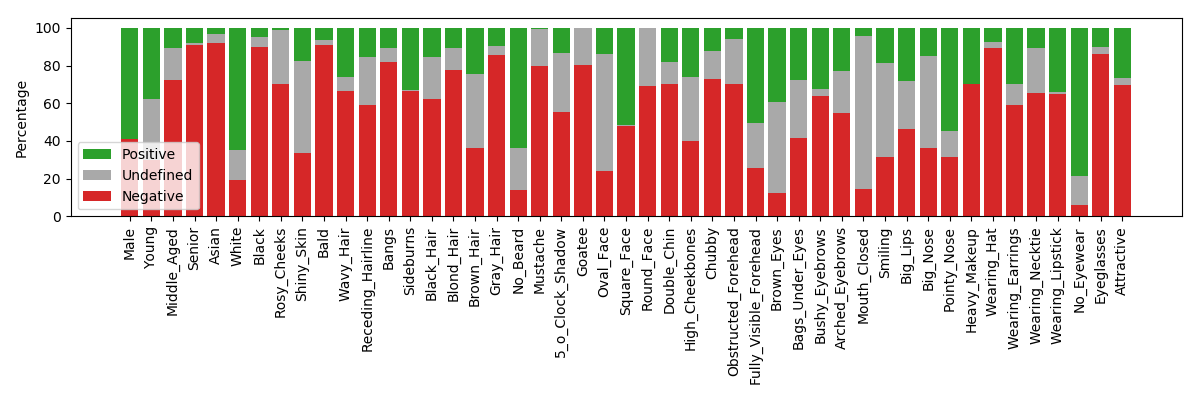}
\caption{Annotation distribution of the proposed MAAD-Face database. For each of the 47 attributes, green indicates the percentage of positive annotations, red indicates the percentage of negatively annotated images, and grey represents the percentage of images that have an undefined annotation for the attribute.}
\label{fig:annotationDistribution}
\end{figure*}

The biggest advantage of MAAD-Face is its large number of high-quality attribute annotations.
Since it builds on the VGGFace2 database, it consists of over 9.1k identities with over 3.3M face images of various poses, ages, and illuminations.
MAAD-Face has annotations for 47 distinctive attributes with a total of 38.3M annotations.
On average $37.5\pm3.7$ annotations are defined per image.
Figure \ref{fig:annotationDistribution} shows the annotation distribution of MAAD-Face for all 47 attributes.
For each attribute, green indicates the percentage of positive annotations, red indicates the percentage of negatively annotated images, and grey represents the percentage of images with undefined annotations.
Some attributes have a low number of positive annotations, such as \textit{Mustache} (16.6k) or \textit{Goatee} (9.2k) and instead, a higher number of undefined annotations.
This way, we can ensure the high correctness of the annotations as explained in Section \ref{sec:AttributeannotationGeneration} (accuracy \textit{Mustache} 98\%, accuracy \textit{Goatee} 95\%).
In total, this leads to MAAD-Face having 23.1\% positive, 56.6\% negative, and 20.3\% undefined annotations.
A list of all attributes with the correctness analysis was already discussed with Table \ref{tab:AttributePerformanceMAADFace} in Section \ref{sec:EvaluatingannotationCorrectness}.
The high quality of the attribute annotations is also observable in Figure \ref{fig:SampleImages}.
There, five random sample images are shown with their corresponding attribute annotations.

%9.1k identities
%3.3M images
%
%47 different attributes
%in total 38.3M attribute annotations[
%
%$37.5\pm3.7$ annotations per image 
%describe all relevant statistics
%
%Attribute: Mustache
%Positive (+1), undefined (0), negative (-1)
%16629,  661569,  2629842
%
%Attribute: Goatee
%Positive (+1), undefined (0), negative (-1)
%9229,  643749,  2655062
%
%--> compare with lfw, number of mustache/goatee annotations + quality?
%
%23.1\% positive annotations
%56.6\% negative annotations
%20.3\% undefined
%
%demographics, skin properties, hair style, colors, beard types, face geometry, periocular, mouth, nose, different accessories, and if the face is perceived as attractive

%Add 2-3 sample images with on the right corresponding annotations

\begin{figure*}

\begin{minipage}[t]{0.22\textwidth}
    \vspace{0pt}
    %\centering
    \includegraphics[width=3.8cm,height=3.8cm]{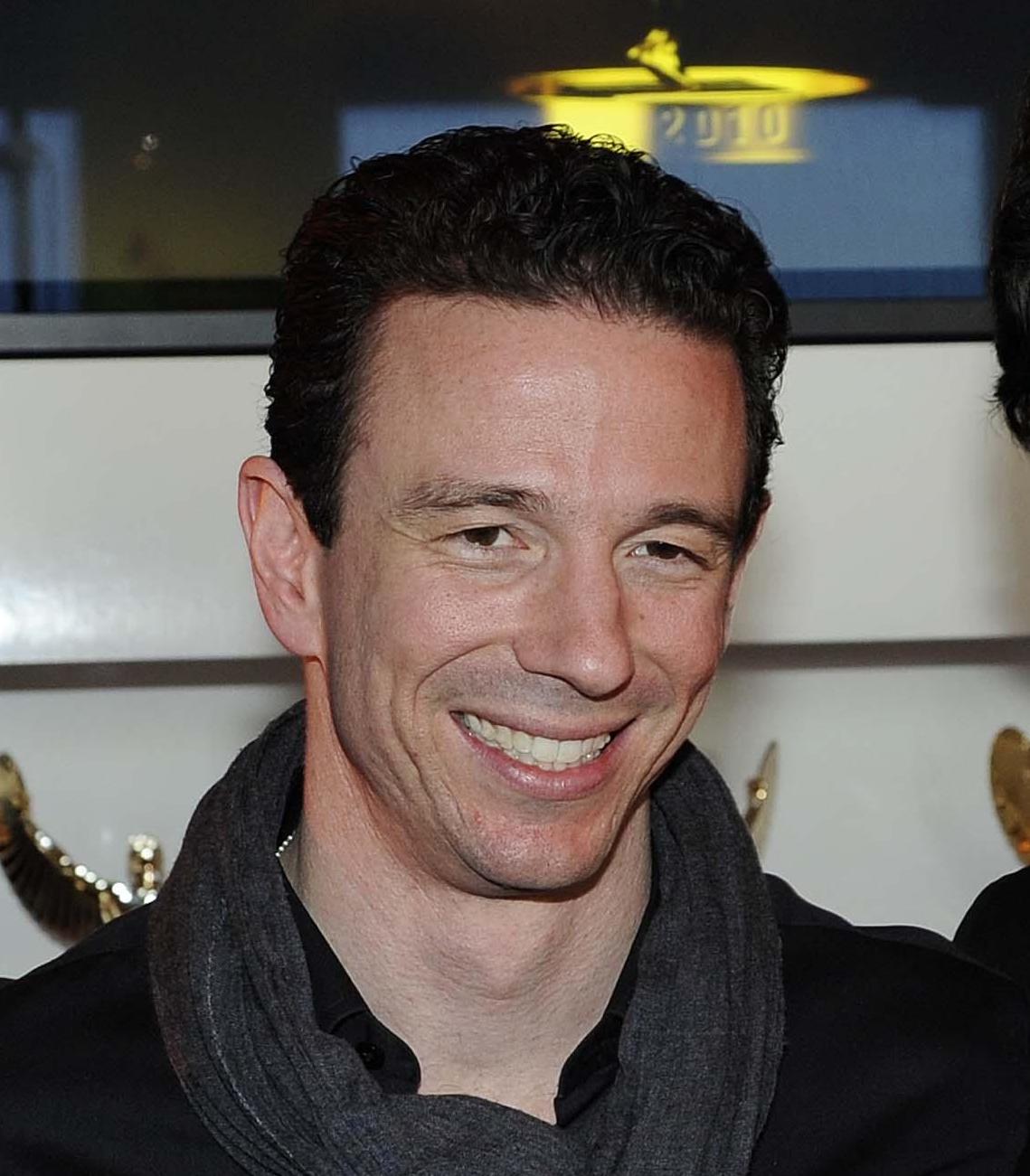}
\end{minipage}
\hfill
\begin{minipage}[t]{0.78\textwidth}
    \vspace{0pt}
    %\centering
    \raisebox{\depth}{
        \footnotesize
        \begin{tabular}{l r l r l r l r}
        \hline
        Male & 1 & Bangs & -1 & Round Face & -1 & Big Lips & 0 \\
        Young & 0 & Sideburns & -1 & Double Chin & -1 & Big Nose & 0 \\
        Middle Aged & 1 & Black Hair & 1 & High Cheekbones & -1 & Pointy Nose & 1 \\
        Senior & -1 & Blond Hair & -1 & Chubby & -1 & Heavy Makeup & -1 \\
        Asian & -1 & Brown Hair & 0 & Obstructed Forehead & 0 & Wearing Hat & -1 \\
        White & 1 & Gray Hair & -1 & Fully Visible Forehead & 0 & Wearing Earrings & -1  \\
        Black & -1 & No Beard & 0 & Brown Eyes & 0 & Wearing Necktie & 1 \\
        Rosy Cheeks & -1 &  Mustache & 0 &  Bags Under Eyes & 1 &  Wearing Lipstick & -1 \\
        Shiny Skin & 0 & 5 o Clock Shadow & 0 & Bushy Eyebrows & 1 & No Eyewear & 1 \\
        Bald & -1 & Goatee & 0 & Arched Eyebrows & -1 & Eyeglasses & -1 \\
        Wavy Hair & -1 & Oval Face & 0 & Mouth Closed & -1 & Attractive & -1 \\
        Receding Hairline & 0 & Square Face & 1 & Smiling & 1 & & \\ \hline
        \end{tabular}
    }
\end{minipage}

\vspace{3mm}

\begin{minipage}[t]{0.22\textwidth}
    \vspace{0pt}
    %\centering
    \includegraphics[width=3.8cm,height=3.8cm]{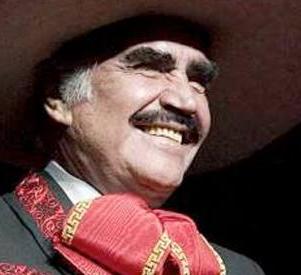}
\end{minipage}
\hfill
\begin{minipage}[t]{0.78\textwidth}
    \vspace{0pt}
    %\centering
    \raisebox{\depth}{
        \footnotesize
        \begin{tabular}{l r l r l r l r}
        \hline
        Male & 1 & Bangs & -1 & Round Face & 0 & Big Lips & 0 \\
        Young & -1 & Sideburns & 1 & Double Chin & 1 & Big Nose & 1 \\
        Middle Aged & -1 & Black Hair & 0 & High Cheekbones & 0 & Pointy Nose & -1 \\
        Senior & 1 & Blond Hair & -1 & Chubby & 1 & Heavy Makeup & -1 \\
        Asian & -1 & Brown Hair & -1 & Obstructed Forehead & 1 & Wearing Hat & 1 \\
        White & 0 & Gray Hair & 1 & Fully Visible Forehead & -1 & Wearing Earrings & -1 \\
        Black & -1 & No Beard & -1 & Brown Eyes & 0 & Wearing Necktie & -1 \\
        Rosy Cheeks & 0 & Mustache & 1 & Bags Under Eyes & 0 & Wearing Lipstick & -1 \\
        Shiny Skin & 1 & 5 o Clock Shadow & -1 & Bushy Eyebrows & 1 & No Eyewear & 1 \\
        Bald & -1 & Goatee & -1 & Arched Eyebrows & -1 & Eyeglasses & -1 \\
        Wavy Hair & -1 & Oval Face & -1 & Mouth Closed & 0 & Attractive & -1 \\
        Receding Hairline & 0 & Square Face & 1 & Smiling & 0 & & \\ \hline
        \end{tabular}
    }
\end{minipage}

%\vspace{3mm}
%
%\begin{minipage}[t]{0.22\textwidth}
%    \vspace{0pt}
%    %\centering
%    \includegraphics[width=3.8cm,height=3.8cm]{imgs/0381_01.jpg}
%\end{minipage}
%\hfill
%\begin{minipage}[t]{0.78\textwidth}
%    \vspace{0pt}
%    %\centering
%    \raisebox{\depth}{
%        \footnotesize
%        \begin{tabular}{l r l r l r l r}
%        \hline
%        Male & 1 & Bangs & -1 & Round Face & -1 & Big Lips & -1 \\
%        Young & -1 & Sideburns & 1 & Double Chin & 1 & Big Nose & 1 \\
%        Middle Aged & 1 & Black Hair & -1 & High Cheekbones & 0 & Pointy Nose & -1 \\
%        Senior & -1 & Blond Hair & -1 & Chubby & 1 & Heavy Makeup & -1 \\
%        Asian & -1 & Brown Hair & -1 & Obstructed Forehead & -1 & Wearing Hat & 1 \\
%        White & 1 & Gray Hair & 1 & Fully Visible Forehead & 1 & Wearing Earrings & -1  \\
%        Black & -1 & No Beard & 0 & Brown Eyes & 0 & Wearing Necktie & 0 \\
%        Rosy Cheeks & 0 &  Mustache & 0 &  Bags Under Eyes & 1 &  Wearing Lipstick & -1 \\
%        Shiny Skin & 0 & 5 o Clock Shadow & 0 & Bushy Eyebrows & 1 & No Eyewear & 1 \\
%        Bald & -1 & Goatee & 0 & Arched Eyebrows & -1 & Eyeglasses & -1 \\
%        Wavy Hair & -1 & Oval Face & -1 & Mouth Closed & 0 & Attractive & -1 \\
%        Receding Hairline & 1 & Square Face & 1 & Smiling & 0 & & \\ \hline
%        \end{tabular}
%    }
%\end{minipage}

\vspace{3mm}
\begin{minipage}[t]{0.22\textwidth}
    \vspace{0pt}
    %\centering
    \includegraphics[width=3.8cm,height=3.8cm]{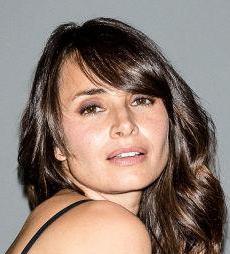}
\end{minipage}
\hfill
\begin{minipage}[t]{0.78\textwidth}
    \vspace{0pt}
    %\centering
    \raisebox{\depth}{
        \footnotesize
        \begin{tabular}{l r l r l r l r}
        \hline
        Male & -1 & Bangs & 1 & Round Face & 0 & Big Lips & 0 \\
        Young & 1 & Sideburns & -1 & Double Chin & -1 & Big Nose & -1 \\
        Middle Aged & -1 & Black Hair & -1 & High Cheekbones & 1 & Pointy Nose & 1 \\
        Senior & -1 & Blond Hair & -1 & Chubby & -1 & Heavy Makeup & 1 \\
        Asian & -1 & Brown Hair & 1 & Obstructed Forehead & 0 & Wearing Hat & -1 \\
        White & 1 & Gray Hair & -1 & Fully Visible Forehead & -1 & Wearing Earrings & 1  \\
        Black & -1 & No Beard & 1 & Brown Eyes & 0 & Wearing Necktie & -1 \\
        Rosy Cheeks & 0 &  Mustache & -1 &  Bags Under Eyes & -1 &  Wearing Lipstick & 1 \\
        Shiny Skin & 0 & 5 o Clock Shadow & -1 & Bushy Eyebrows & -1 & No Eyewear & 1 \\
        Bald & -1 & Goatee & -1 & Arched Eyebrows & -1 & Eyeglasses & -1 \\
        Wavy Hair & 1 & Oval Face & 0 & Mouth Closed & 0 & Attractive & 1 \\
        Receding Hairline & -1 & Square Face & -1 & Smiling & 0 & & \\ \hline
        \end{tabular}
    }
\end{minipage}

\vspace{3mm}

\begin{minipage}[t]{0.22\textwidth}
    \vspace{0pt}
    %\centering
    \includegraphics[width=3.8cm,height=3.8cm]{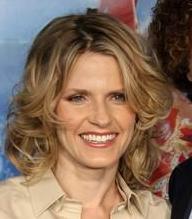}
\end{minipage}
\hfill
\begin{minipage}[t]{0.78\textwidth}
    \vspace{0pt}
    %\centering
    \raisebox{\depth}{
        \footnotesize
        \begin{tabular}{l r l r l r l r}
        \hline
        Male & -1 & Bangs & -1 & Round Face & -1 & Big Lips & -1 \\
        Young & 1 & Sideburns & -1 & Double Chin & -1 & Big Nose & -1 \\
        Middle Aged & -1 & Black Hair & 0 & High Cheekbones & 1 & Pointy Nose & 1 \\
        Senior & -1 & Blond Hair & 0 & Chubby & -1 & Heavy Makeup & 1 \\
        Asian & -1 & Brown Hair & 0 & Obstructed Forehead & -1 & Wearing Hat & -1 \\
        White & 1 & Gray Hair & 0 & Fully Visible Forehead & 1 & Wearing Earrings & 1  \\
        Black & -1 & No Beard & 1 & Brown Eyes & -1 & Wearing Necktie & -1 \\
        Rosy Cheeks & 0 &  Mustache & -1 &  Bags Under Eyes & -1 &  Wearing Lipstick & 1 \\
        Shiny Skin & 0 & 5 o Clock Shadow & -1 & Bushy Eyebrows & -1 & No Eyewear & 1 \\
        Bald & -1 & Goatee & -1 & Arched Eyebrows & 0 & Eyeglasses & -1 \\
        Wavy Hair & 1 & Oval Face & 1 & Mouth Closed & 0 & Attractive & 1 \\
        Receding Hairline & -1 & Square Face & -1 & Smiling & 1 & & \\ \hline
        \end{tabular}
    }
\end{minipage}

\vspace{3mm}

\begin{minipage}[t]{0.22\textwidth}
    \vspace{0pt}
    %\centering
    \includegraphics[width=3.8cm,height=3.8cm]{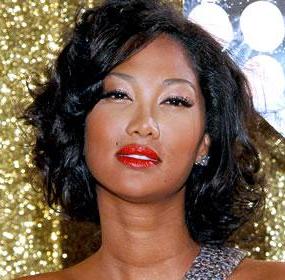}
\end{minipage}
\hfill
\begin{minipage}[t]{0.78\textwidth}
    \vspace{0pt}
    %\centering
    \raisebox{\depth}{
        \footnotesize
        \begin{tabular}{l r l r l r l r}
        \hline
        Male & -1 & Bangs & -1 & Round Face & 0 & Big Lips & 1 \\
        Young & 0 & Sideburns & -1 & Double Chin & -1 & Big Nose & -1 \\
        Middle Aged & 1 & Black Hair & 1 & High Cheekbones & 1 & Pointy Nose & 0 \\
        Senior & -1 & Blond Hair & -1 & Chubby & -1 & Heavy Makeup & 1 \\
        Asian & 1 & Brown Hair & 0 & Obstructed Forehead & -1 & Wearing Hat & -1 \\
        White & -1 & Gray Hair & -1 & Fully Visible Forehead & 1 & Wearing Earrings & 1 \\
        Black & -1 & No Beard & 1 & Brown Eyes & 1 & Wearing Necktie & -1 \\
        Rosy Cheeks & -1 & Mustache & -1 & Bags Under Eyes & -1 & Wearing Lipstick & 1 \\
        Shiny Skin & 1 & 5 o Clock Shadow & -1 & Bushy Eyebrows & -1 & No Eyewear & 0 \\
        Bald & -1 & Goatee & -1 & Arched Eyebrows & 1 & Eyeglasses & -1 \\
        Wavy Hair & 1 & Oval Face & 0 & Mouth Closed & -1 & Attractive & 1 \\
        Receding Hairline & -1 & Square Face & -1 & Smiling & -1 & & \\ \hline
        \end{tabular}
    }
\end{minipage}

\caption{Samples images from MAAD-Face with the corresponding 47 attribute-annotations.}
\label{fig:SampleImages}
\end{figure*}

%\rule{0pt}{3ex}\noindent
%\parbox[t]{2mm}{\multirow{4}{*}{\rotatebox[origin=c]{90}{Environment}}}

\begin{table*}[]
\renewcommand{\arraystretch}{1.2}
\centering
\caption{Attribute annotation analysis of MAAD-Face based on the ground truth of three human evaluators. The annotation quality is reported in terms of accuracy, precision, and recall. Main source describe from which dataset most of the annotations are transferred from.}
\label{tab:AttributePerformanceMAADFace}
\begin{tabular}{llllrrr}
\Xhline{2\arrayrulewidth}
Main source & Category      & Class           & Attribute                    & Accuracy  & Precision & Recall \\
\hline
CelebA        & Demographics  & Gender              & Male                     & 0.99 & 0.98      & 1.00   \\
CelebA        &               & Age                 & Young                    & 0.99 & 1.00      & 0.98   \\
LFW           &               &                     & Middle Aged             & 0.93 & 0.98      & 0.89   \\
LFW           &               &                     & Senior                   & 0.97 & 0.96      & 0.98   \\
LFW           &               & Race                & Asian                    & 0.90 & 0.88      & 0.92   \\
LFW           &               &                     & White                    & 0.89 & 1.00      & 0.82   \\
LFW           &               &                     & Black                    & 0.94 & 0.90      & 0.98   \\
CelebA        & Skin          & Rosy Cheeks        & Rosy Cheeks             & 0.99 & 0.98      & 1.00   \\
LFW           &               & Shiny Skin         & Shiny Skin              & 0.77 & 0.84      & 0.74   \\
CelebA        & Hair          & Hairstyle           & Bald                     & 0.96 & 0.92      & 1.00   \\
CelebA        &               &                     & Wavy Hair               & 0.99 & 1.00      & 0.98   \\
CelebA        &               & Receding Hairline  & Receding Hairline       & 0.77 & 0.54      & 1.00     \\
CelebA        &               & Bangs               & Bangs                    & 0.98 & 0.96      & 1.00   \\
CelebA        &               & Sideburns           & Sideburns                & 0.93 & 0.88      & 0.98   \\
CelebA        &               & Haircolor           & Black Hair              & 0.98 & 0.96      & 1.00   \\
CelebA        &               &                     & Blond Hair              & 1.00 & 1.00      & 1.00   \\
CelebA        &               &                     & Brown Hair              & 0.97 & 0.94      & 1.00   \\
CelebA        &               &                     & Gray Hair               & 0.95 & 0.90      & 1.00   \\
CelebA        & Beard         & Beard               & No Beard                & 0.98 & 1.00      & 0.96   \\
CelebA        &               &                     & Mustache                 & 0.98 & 0.98      & 0.98   \\
CelebA        &               &                     & 5 o Clock Shadow      & 0.97 & 0.94      & 1.00   \\
CelebA        &               &                     & Goatee                   & 0.95 & 0.90      & 1.00   \\
LFW           & Face Geometry & Face Shape          & Oval Face               & 0.81 & 0.90      & 0.76   \\
LFW           &               &                     & Square Face             & 0.80 & 0.78      & 0.81   \\
LFW           &               &                     & Round Face              & 0.69 & 0.56      & 0.76   \\
CelebA        &               & Double Chin        & Double Chin             & 0.94 & 0.88      & 1.00   \\
CelebA        &               & High Cheekbones    & High Cheekbones         & 0.92 & 0.92      & 0.92   \\
CelebA        &               & Chubby              & Chubby                   & 0.94 & 0.88      & 1.00   \\
LFW           &               & Forehead visibility & Obstructed Forehead     & 0.91 & 0.94      & 0.89   \\
LFW           &               &                     & Fully Visible Forehead & 0.80 & 0.75      & 1.00   \\
LFW           & Periocular    & Brown Eyes         & Brown Eyes              & 0.68 & 0.44      & 0.85   \\
LFW           &               & Bags Under Eyes   & Bags Under Eyes        & 0.68 & 0.40      & 0.91   \\
CelebA        &               & Bushy Eyebrows     & Bushy Eyebrows          & 0.95 & 0.94      & 0.96   \\
CelebA        &               & Arched Eyebrows    & Arched Eyebrows         & 1.00 & 1.00      & 1.00   \\
LFW           & Mouth         & Mouth Closed          & Mouth Closed            & 0.84 & 0.80      & 0.87   \\
CelebA        &               & Smiling             & Smiling                  & 0.95 & 1.00      & 0.91   \\
LFW           &               & Big Lips           & Big Lips                & 0.70 & 0.50      & 0.83   \\
CelebA        & Nose          & Nose type           & Big Nose                & 0.97 & 0.98      & 0.96   \\
LFW           &               &                     & Pointy Nose             & 0.88 & 0.88      & 0.88   \\
CelebA        & Accessories   & Heavy Makeup       & Heavy Makeup            & 0.98 & 0.98      & 0.98   \\
CelebA        &               & Wearing Hat        & Wearing Hat             & 0.92 & 0.84      & 1.00   \\
CelebA        &               & Wearing Earrings   & Wearing Earrings        & 0.83 & 0.70      & 0.95   \\
LFW           &               & Wearing Necktie    & Wearing Necktie         & 0.91 & 0.84      & 0.98   \\
CelebA        &               & Wearing Lipstick   & Wearing Lipstick        & 0.95 & 0.90      & 1.00   \\
LFW           &               & Eyeglasses          & No Eyewear              & 0.98 & 0.98      & 0.98   \\
CelebA        &               &                     & Eyeglasses               & 0.90 & 0.80      & 1.00   \\
CelebA        & Other         & Attractive          & Attractive               & 1.00 & 1.00      & 1.00   \\
\hline
              &               &  Total                   &                          & 0.91 & 0.87      & 0.94  \\
              \Xhline{2\arrayrulewidth} 
\end{tabular}
\end{table*}

\begin{figure}
%\captionsetup[subfloat]{farskip=5pt,captionskip=1pt}
\centering    
\subfloat[]{%
       \includegraphics[width=0.08\textwidth]{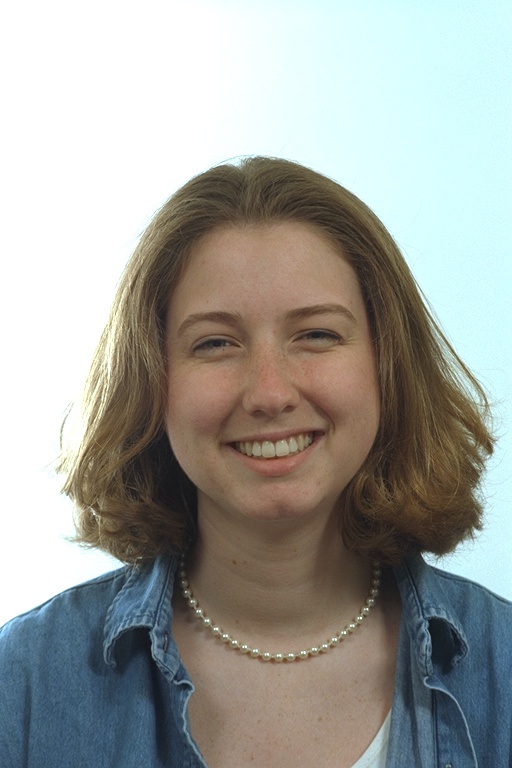}}  
\subfloat[]{%
       \includegraphics[width=0.08\textwidth]{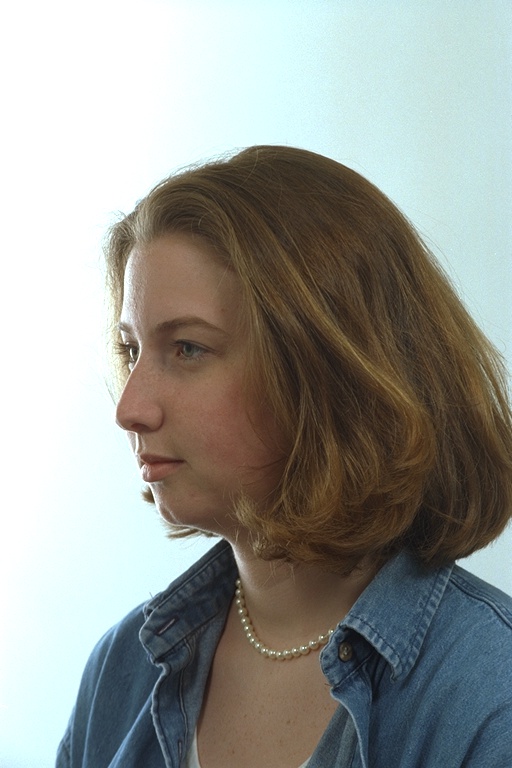}}
\subfloat[]{%
       \includegraphics[width=0.08\textwidth]{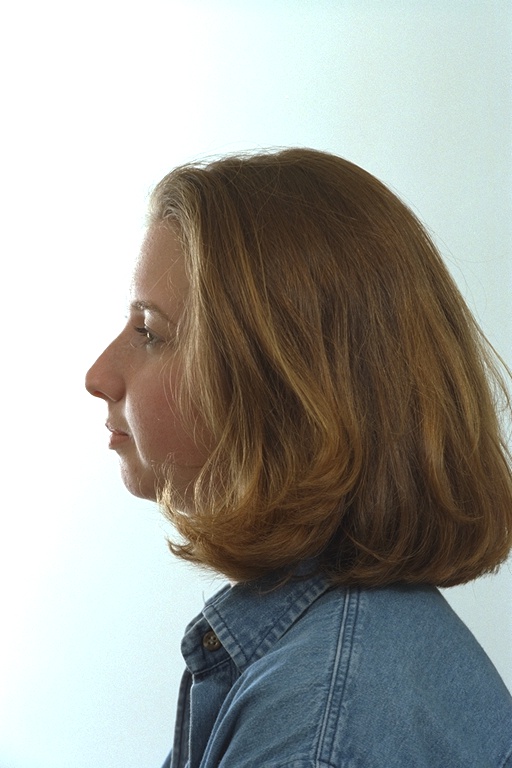}}
\subfloat[]{%
       \includegraphics[width=0.08\textwidth]{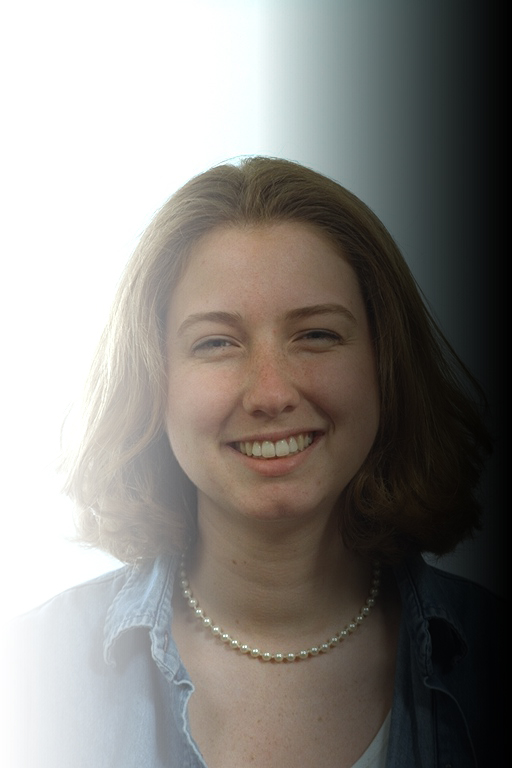}}  
\subfloat[]{%
       \includegraphics[width=0.08\textwidth]{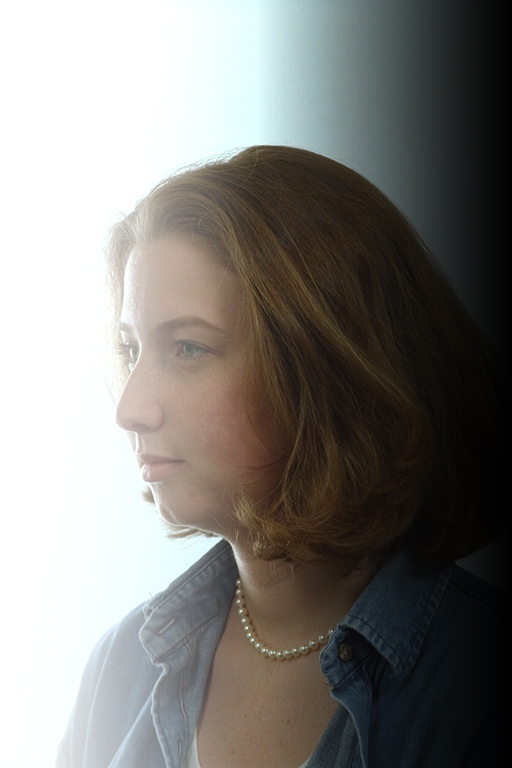}}
\subfloat[]{%
       \includegraphics[width=0.08\textwidth]{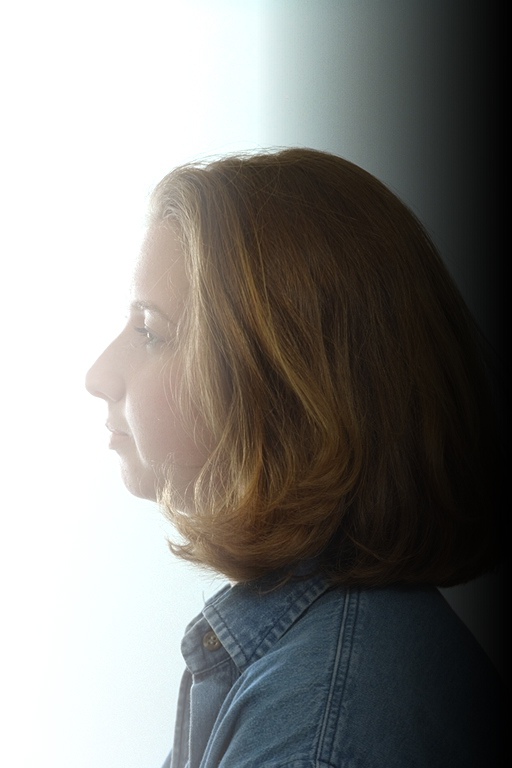}}

\caption{Sample Images \cite{ColorFERET} for the pose and lighting evaluation. Three different head poses are considered from frontal to profile images. To also analyse the effect of lighting, artificial lighting was introduced (see Figures d-f).
} 
\label{fig:SampleImages}
\end{figure}

\subsection{Evaluating MAC Performance under Pose and Lighting}
\label{sec:MAC_Pose_Lighting}
In this section, we evaluate the performance of the MAC against the industry product Microsoft Azure\footnote{\url{https://azure.microsoft.com/en-us/services/cognitive-services/face}} under different headposes and lighting conditions. 
The experiment was conducted on the ColorFeret database \cite{ColorFERET} consisting of over 11k images with different poses such as frontal (2.7k), profile (2.7k) and head poses in between (5.9k).
Figure \ref{fig:SampleImages} shows some sample images visualizing the head poses and lighting conditions.
While our MAC classifier is able to accurately state 47 different attributes, we focus on the attributes that Microsoft Azure and the ColorFeret dataset have in common. 
This allows a comparison of our approach with an industry product under several conditions.

Table \ref{tab:MAC_SOTA} shows the described analysis. 
In general, our MAC classifier reaches high performances on most attributes and additionally turns out to be relatively robust against the changes in the head poses as well as the lighting.
The weakest performance is observed for the only vaguely-defined age classes.
Similar to our MAC approach, Microsoft Azure turns out to have accurate and robust predictions under both lighting conditions. However, Microsoft Azure tends to reject many images when it comes to non-frontal poses or difficult lighting conditions.
Under optimal lighting conditions, 85\% of the profile images are rejected and in combination with the more challenging lighting, over 99\% of the images are rejected without any predictions.
This short analysis showed that our MAC classifier used for the proposed annotation transfer pipeline is well suited and reaches comparable performances to the industry product Microsoft Azure.

\subsection{Evaluating Annotation-Correctness}
\label{sec:EvaluatingannotationCorrectness}

In this section, we evaluate the quality of attribute annotations from three face datasets, LFW, CelebA, and MAAD-Face.
The quality refers to the correctness of the annotations compared to the annotations of human evaluators.
%We will demonstrate that the quality of the attribute annotations of MAAD-Face is significantly stronger than LFW and CelebA, despite having 137 and 15 times more annotations.
The annotation-correctness of each attribute in LFW, CelebA, and MAAD-Face was manually evaluated by three human evaluators.
For each attribute, the evaluators got 50 positively-annotated and 50 negatively-annotated images to prevent a bias evaluation due to unbalanced testing data.
Since the choice of the images submitted to the evaluators affect the correctness evaluation, the selection of these images is based on a randomized process to prevent a human-bias in the choice of these images.
%\textcolor{blue}{These were randomly chosen such that no image is seen by more than one evaluator.
%This aims at preventing the inter-annotator problem described in \cite{DBLP:conf/cvpr/BenensonPF19, jansen, DBLP:conf/mir/NowakR10}.}
Then, each evaluator was asked to carefully annotate these images for the given attribute.
This led to over 16k manually created annotations\footnote{Please note that this only represents a small fraction of all annotations and additionally reflects the subjective opinion of the three evaluators.
Therefore, the results should not be considered as absolute values but should rather be used as indicators.}.
The manually created annotations are used to compute the accuracy, precision, and recall for each attribute of the database.
The accuracy refers to the percentage of correct annotations, where the ground truth is determined by the human evaluators.
Precision is defined as the number of true positives over the number of true and false positives In our context, precision refers to "What proportion of positive annotated samples in the database is also positively annotated by the human evaluators?".
Recall is defined as the number of true positives over the number of true positives and false negatives.
In our context, recall refers to "What proportion of positive human annotations are identified correctly?".
Tables \ref{tab:AttributePerformanceLFW}, \ref{tab:AttributePerformanceCelebA}, and \ref{tab:AttributePerformanceMAADFace} present the results for this analysis on LFW, CelebA, and MAAD-Face.

\begin{table*}[]
\renewcommand{\arraystretch}{1.2}
\centering
\caption{
Analysis of the attribute annotation correctness. The correctness of the attribute annotations is shown for the most relevant databases, LFW, CelebA, and MAAD-Face, since these contain annotations for a high number of distinct attributes. 
The correctness was evaluated by three human evaluators.
In total, MAAD-Face does not only provide the highest number of attribute annotations, it additionally provides annotations of much higher quality than related databases.}
\label{tab:DatabaseCorrectness}
\begin{tabular}{lrrrrrrr}
\Xhline{2\arrayrulewidth}
                &                    &                  & \multicolumn{2}{c}{Attribute annotations}                                                    & \multicolumn{3}{c}{Attribute annotation correctness}  \\
                \cmidrule(rl){4-5} \cmidrule(rl){6-8}
Database        & Num. of subjects & Num. of images & Distinctive attributes & Number of annotations & Accuracy & Precision & Recall \\
\hline
LFW             & 5.7k               & 13.2k              & 74                     & 0.9M             & 0.72           & 0.61            & 0.84         \\
CelebA          & 10.0k                & 0.2M             & 40                     & 8.0M             & 0.85           & 0.83            & 0.89         \\
MAAD-Face (this paper) & 9.1k               & 3.3M             & 47                     & 123.9M            & \textbf{0.91}           & \textbf{0.87}            & \textbf{0.94}      \\
\Xhline{2\arrayrulewidth}  
\end{tabular}
\end{table*}

\subsubsection{LFW}
For LFW (Table \ref{tab:AttributePerformanceLFW}), many attributes show a very weak performance and thus, a low correlation with the annotations of the human evaluators.
Young age group annotations (baby, child, youth) are close to a random accuracy and additionally often have a small precision.
This is also observed e.g. for \textit{frowning, chubby, curly hair, wavy hair, bangs, goatee, and square face}.
Moreover, annotations for \textit{attractive man} are mostly placed on female faces.
In general, there is a big mismatch between the annotations of LFW and the annotations of the human evaluators.
The accuracy for most attributes is below 80\% and only 5 out of 76 attributes have an accuracy of over 90\%.
Over all attributes, this leads to an accuracy of 72\%, a precision of 61\%, and a recall of 84\%.
The high gap between the low precision and the relatively high recall indicates that there are a lot of false-positive annotations in LFW.

\subsubsection*{CelebA}
The attribute performance for CelebA is shown in Table \ref{tab:AttributePerformanceCelebA}.
It has annotations for 40 binary attribute, which is a lower number than on LFW.
However, these annotations are of much higher quality.
Only 2 attributes have an accuracy of less than 70\% and 14 attributes even reach over 90\%.
Over all attributes, the accuracy is 85\%, the precision is 83\%, and the recall is 89\%.
Similar to LFW, there is a tendency that most of the wrong annotations are within the positives.

\subsubsection*{MAAD-Face}
Table \ref{tab:AttributePerformanceMAADFace} shows the attribute performance of MAAD-Face.
MAAD-Face has 47 binary attributes.
In the evaluation against the human annotations, 3 attributes reach a performance of below 70\%.
However, also 34 attributes reach over 90\% accuracy with the majority of close to 100\%.
Over all attributes, this leads to an accuracy of 91\%, a precision of 87\%, and a recall of 94\%.

\subsubsection*{Summary}
Table \ref{tab:DatabaseCorrectness} shows the properties of the investigated databases including the overall performance of our annotation-correctness study.
%On all databases, it is observed that the precision is always a bit lower than the recall.
%This might indicate that (a) evaluators tend to negative, or (b) databases prefer positive class
Even though LFW provides the highest number of binary attributes, it provides the lowest number of attribute annotations with the lowest annotation qualities.
Only 72\% of the investigated annotations match the annotations of the human evaluators.
CelebA consists of 8.0M attribute annotations of 40 binary attributes.
Moreover, with an accuracy of 85\%, the quality of these annotations is significantly higher.
In terms of numbers of annotations and annotation-quality, MAAD-Face exceeds the other databases.
It provides 47 binary attributes with a total of 123.9M annotations.
This is 15 times higher than CelebA and 137 times higher than LFW.
Moreover, the annotations quality (in terms of accuracy, precision, and recall) is significantly higher than the other databases.
91\% of the MAAD-Face annotations match the annotations of the human evaluators.
Consequently, MAAD-Face provides significantly more and higher-quality attribute annotations.

\begin{table*}[]
\renewcommand{\arraystretch}{1.2}
\centering
\caption{Attribute annotation analysis of LFW based on the ground truth of three human evaluators. The annotation quality is reported in terms of accuracy, precision, and recall.}
\label{tab:AttributePerformanceLFW}
\begin{tabular}{cc}
\begin{tabu}{llll}
\Xhline{2\arrayrulewidth}
Attribute                        & Acc  & Precision & Recall \\
\hline
Male                         & 0.89 & 0.96      & 0.84   \\
Asian                        & 0.86 & 0.74      & 0.97   \\
White                        & 0.74 & 0.98      & 0.66   \\
Black                        & 0.91 & 0.84      & 0.98   \\
Baby                         & 0.54 & 0.08      & 1.00   \\
Child                        & 0.55 & 0.10      & 1.00   \\
Youth                        & 0.56 & 0.14      & 0.88   \\
Middle Aged                 & 0.67 & 0.90      & 0.62   \\
Senior                       & 0.87 & 0.94      & 0.82   \\
Black Hair                  & 0.78 & 0.88      & 0.73   \\
Blond Hair                  & 0.91 & 0.84      & 0.98   \\
Brown Hair                  & 0.70 & 0.60      & 0.75   \\
Bald                         & 0.74 & 0.50      & 0.96   \\
No Eyewear                  & 0.91 & 0.98      & 0.86   \\
Eyeglasses                   & 0.91 & 0.88      & 0.94   \\
Sunglasses                   & 0.86 & 0.72      & 1.00   \\
Moustache                     & 0.84 & 0.72      & 0.95   \\
Smiling                      & 0.87 & 0.80      & 0.93   \\
Frowning                     & 0.61 & 0.22      & 1.00   \\
Chubby                       & 0.53 & 0.16      & 0.62   \\
Blurry                       & 0.69 & 0.90      & 0.63   \\
Harsh Lighting              & 0.64 & 0.92      & 0.59   \\
Flash                        & 0.73 & 0.66      & 0.77   \\
Soft Lighting               & 0.75 & 0.66      & 0.80   \\
Outdoor                      & 0.83 & 0.82      & 0.84   \\
Curly Hair                  & 0.51 & 0.02      & 1.00   \\
Wavy Hair                   & 0.50 & 0.08      & 0.50   \\
Straight Hair               & 0.60 & 0.78      & 0.54   \\
Receding Hairline           & 0.75 & 0.62      & 0.84   \\
Bangs                        & 0.54 & 0.08      & 1.00   \\
Sideburns                    & 0.61 & 0.40      & 0.69   \\
Fully Visible Forehead     & 0.79 & 1.00      & 0.70   \\
Partially Visible Forehead & 0.82 & 0.80      & 0.83   \\
Obstructed Forehead         & 0.62 & 0.24      & 1.00   \\
Bushy Eyebrows              & 0.64 & 0.42      & 0.75   \\
Arched Eyebrows             & 0.79 & 0.80      & 0.78   \\
Narrow Eyes                 & 0.69 & 0.46      & 0.85  \\
\Xhline{2\arrayrulewidth}
\end{tabu}
&
\begin{tabu}{llll}
\Xhline{2\arrayrulewidth}
Attribute                     & Acc  & Precision & Recall \\
\hline
Eyes Open                & 0.73 & 0.96      & 0.66   \\
Big Nose                 & 0.75 & 0.54      & 0.93   \\
Pointy Nose              & 0.80 & 0.82      & 0.79   \\
Big Lips                 & 0.73 & 0.56      & 0.85   \\
Mouth Closed             & 0.86 & 0.82      & 0.89   \\
Mouth Slightly Open     & 0.79 & 0.88      & 0.75   \\
Mouth Wide Open         & 0.93 & 0.88      & 0.98   \\
Teeth Not Visible       & 0.86 & 0.78      & 0.93   \\
No Beard                 & 0.69 & 1.00      & 0.62   \\
Goatee                    & 0.62 & 0.24      & 1.00   \\
Round Jaw                & 0.77 & 0.76      & 0.78   \\
Double Chin              & 0.66 & 0.34      & 0.94   \\
Wearing Hat              & 0.69 & 0.40      & 0.95   \\
Oval Face                & 0.59 & 0.78      & 0.57   \\
Square Face              & 0.55 & 0.12      & 0.86   \\
Round Face               & 0.81 & 0.72      & 0.88   \\
Color Photo              & 0.57 & 1.00      & 0.54   \\
Posed Photo              & 0.64 & 0.32      & 0.89   \\
Attractive Man           & 0.62 & 0.26      & 0.93   \\
Attractive Woman         & 0.75 & 0.50      & 1.00   \\
Indian                    & 0.65 & 0.32      & 0.94   \\
Gray Hair                & 0.89 & 0.94      & 0.85   \\
Bags Under Eyes         & 0.75 & 0.76      & 0.75   \\
Heavy Makeup             & 0.88 & 0.76      & 1.00   \\
Rosy Cheeks              & 0.63 & 0.30      & 0.88   \\
Shiny Skin               & 0.66 & 0.44      & 0.79   \\
Pale Skin                & 0.82 & 0.90      & 0.78   \\
5 o Clock Shadow       & 0.59 & 0.18      & 1.00   \\
Strong Nose-Mouth Lines & 0.86 & 0.88      & 0.85   \\
Wearing Lipstick         & 0.81 & 0.64      & 0.97   \\
Flushed Face             & 0.61 & 0.28      & 0.82   \\
High Cheekbones          & 0.81 & 0.70      & 0.90   \\
Brown Eyes               & 0.44 & 0.46      & 0.44   \\
Wearing Earrings         & 0.79 & 0.58      & 1.00   \\
Wearing Necktie          & 0.76 & 0.66      & 0.83   \\
Wearing Necklace         & 0.61 & 0.22      & 1.00   \\
\hline
Total                     & 0.72 & 0.61      & 0.84  \\
\Xhline{2\arrayrulewidth}
\end{tabu} 
\end{tabular}
\end{table*}

\begin{table}[]
\renewcommand{\arraystretch}{1.2}
\centering
\caption{Attribute annotation analysis of CelebA based on the ground truth of three human evaluators. The annotation quality is reported in terms of accuracy, precision, and recall.}
\label{tab:AttributePerformanceCelebA}
\begin{tabular}{llll}
\Xhline{2\arrayrulewidth}
Attribute                 & Acc  & Precision & Recall \\
\hline
5 o Clock Shadow   & 0.85 & 0.74      & 0.95   \\
Arched Eyebrows      & 0.89 & 0.92      & 0.87   \\
Attractive            & 0.81 & 0.74      & 0.86   \\
Bags Under Eyes     & 0.80 & 0.80      & 0.80   \\
Bald                  & 0.84 & 0.68      & 1.00   \\
Bangs                 & 0.75 & 0.50      & 1.00   \\
Big Lips             & 0.73 & 0.84      & 0.69   \\
Big Nose             & 0.79 & 0.86      & 0.75   \\
Black Hair           & 0.87 & 0.96      & 0.81   \\
Blond Hair           & 0.94 & 0.94      & 0.94   \\
Blurry                & 0.88 & 0.78      & 0.98   \\
Brown Hair           & 0.90 & 0.88      & 0.92   \\
Bushy Eyebrows       & 0.81 & 0.78      & 0.83   \\
Chubby                & 0.83 & 0.66      & 1.00   \\
Double Chin          & 0.76 & 0.58      & 0.91   \\
Eyeglasses            & 0.96 & 0.92      & 1.00   \\
Goatee                & 0.93 & 0.94      & 0.92   \\
Gray Hair            & 0.98 & 0.98      & 0.98   \\
Heavy Makeup         & 0.90 & 0.92      & 0.88   \\
High Cheekbones      & 0.88 & 0.86      & 0.90   \\
Male                  & 1.00 & 1.00      & 1.00   \\
Mouth Slightly Open & 0.90 & 0.88      & 0.92   \\
Mustache              & 0.95 & 0.94      & 0.96   \\
Narrow Eyes          & 0.86 & 0.82      & 0.89   \\
No Beard             & 0.91 & 1.00      & 0.85   \\
Oval Face            & 0.62 & 0.92      & 0.58   \\
Pale Skin            & 0.85 & 0.92      & 0.81   \\
Pointy Nose          & 0.83 & 0.94      & 0.77   \\
Receding Hairline    & 0.66 & 0.38      & 0.86   \\
Rosy Cheeks          & 0.78 & 0.70      & 0.83   \\
Sideburns             & 0.84 & 0.88      & 0.81   \\
Smiling               & 0.94 & 0.92      & 0.96   \\
Straight Hair        & 0.83 & 1.00      & 0.75   \\
Wavy Hair            & 0.82 & 0.66      & 0.97   \\
Wearing Earrings     & 0.93 & 0.88      & 0.98   \\
Wearing Hat          & 1.00 & 1.00      & 1.00   \\
Wearing Lipstick     & 0.91 & 0.90      & 0.92   \\
Wearing Necklace     & 0.86 & 0.80      & 0.91   \\
Wearing Necktie      & 0.85 & 0.72      & 0.97   \\
Young                 & 0.75 & 0.52      & 0.96   \\
\hline
Total                 & 0.85 & 0.83      & 0.89  \\
\Xhline{2\arrayrulewidth}
\end{tabular}
\end{table}

\section{Soft-Biometric based Identity Recognition}
\label{sec:SoftBiometricRecognition}

%Soft-biometric attributes allow recognizing individuals in a privacy-preserving manner.
In this section, we evaluate the discrimination strength of soft-biometric attributes for identity verification and identification.
The use of these attribute might be especially interesting for identity recognition applications with short time windows between the reference and probe images, such as person re-identification.
In the following of this section, we describe the experimental setup.
Afterwards, the results are presented, discussed, and summarized.

\subsection{Experimental Setup}
The high number of face annotations with sufficient quality in MAAD-Face allow us to investigate the usefulness of soft-biometrics for face recognition.
For the experiments, the MAAD-Face database is divided into a 20\% training and a 80\% test set in a subject exclusive manner.
As a result, the training set contains around 630k samples while the test set contains around 2.5M instances. 
This test/train division allows to make the use of a large test set while the training set size is still suitable for training a linear (logistic regression) model \cite{10.5555/1162264}.

For the identification experiments, the test set is further divided into a reference and a probe set.
From each identity, one sample with the most annotated attributes is placed in the reference.
All the others are placed in the probe set.
The identification performance is reported in terms of Cumulative Match Characteristic (CMC) curves \cite{6712705}.
It measures  the identification performance based on the relative ordering of match scores corresponding to each biometric sample in a closed-set identification scenario. 
%It shows how much of the database have to be search (penetration rate) to have a certain probability (hitrate) to find the search identity.
For the verification experiments, all samples pairs are considered.
The face verification performance is reported in terms of false non-match rate (FNMR) at a fixed false match rate (FMR).
These measures are specified for biometric verification evaluation in the international standard \cite{ISO_Metrik}.

\begin{figure}
\centering
\includegraphics[width=0.3\textwidth]{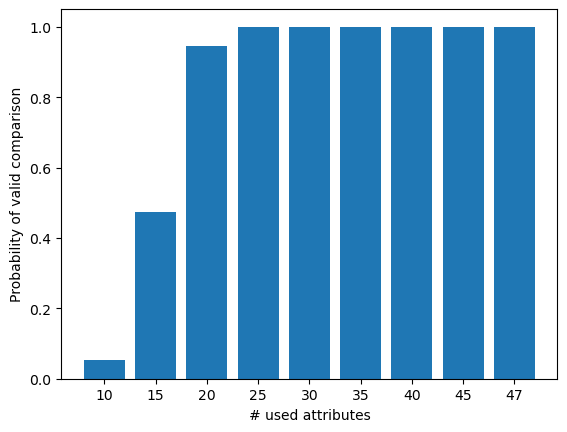}
\caption{Probability that a comparison is valid depending on the number of (most important) attributes used for the comparison. A comparison is considered as valid if at least 10 attributes are annotated in both, the probe and the reference sample.}
\label{fig:ValidRanking}
\end{figure}

To ensure that the comparison of two samples will contain a sufficient number of attributes that jointly appear in both samples, we neglect comparisons with less than 10 overlapping attribute annotations (not valid).
We set this constraint of at least 10 overlapping attributes since we see a direct relation between the number of considered attributes and the expected accuracy of the made decision.
For a low number of overlapping attribute annotations, the decision is made based on less information and thus, a false decision is more likely.
For a high number of overlaps, the decision takes into account more information and therefore, it is more likely that the decision will be correct.
Figure \ref{fig:ValidRanking} shows the probability that a comparison is valid depending on the number of used (most important) attributes.
It can be seen that if a comparison is made using 20 or more of the most important attributes, the probability that the comparison is neglected is very low.

The comparison of two samples is made with a joint feature representation.
Therefore, a joint (soft-biometric) feature representation
\begin{align}
x(x_{ref}, x_{probe}) = [x_1^{a_1}, x_2^{a_1}, x_3^{a_1},  x_1^{a_2}, x_2^{a_2}, x_3^{a_2}, \dots]
\end{align}
is computed and a hamming-based and a logistic regression model is utilized for the comparison process itself.
The joint feature representation for the attribute $a$
\begin{align}
x_i^{a}(x_{ref}, x_{probe})= \begin{cases}
1 \quad \mbox{if} \,\, i=1 \,\, \& \,\, x_{ref}^a = x_{probe}^a = \textit{True} \\
1 \quad \mbox{if} \,\, i=2 \,\, \& \,\, x_{ref}^a = x_{probe}^a = \textit{False} \\
1 \quad \mbox{if} \,\, i=3 \,\, \& \,\, x_{ref}^a \neq x_{probe}^a  \\
0 \quad \mbox{otherwise}
\end{cases}
\end{align}
of a reference sample $x_{ref}$ and a probe sample $x_{probe}$ is defined binary by the relation of $a$ depending if the attribute of both samples is annotated as both as True ($x_1^a$), both as False ($x_2^a$), or differently ($x_3^a$).
We chose this kind of representation to ensure that the comparison models can additionally learn the relation between different attributes.
Two simple comparison models are used for the experiments that exploit the joint feature representation.
The first one is a hamming-based model that simply determines the number of equally-annotated attributes in a normalized manner.
The comparison score of this model is given by
\begin{align}
s(x) = 1 - \textit{NHD}(x), 
\end{align}
where \textit{NHD} counts the number of 1's in $x$ and divides it by the number of attributes $|\mathcal{A}|$.
The second one makes use of the training set and trains a logistic regression model on the joint feature representations.
The choice of a simple linear model prevents overfitting and additionally allows to determine the importance of each soft-biometric attribute.

%•	1- NHD (normalized HD)
%•	Min 10 Attribute überlappung
%•	Testset/reference: pro Identität ein sample mit dem meisten attribute annotations
%o	Vergleich gegen alle anderen samples
%
%
%come closer to the answer of the question if the identity is just the sum of soft-biometrics

\subsection{Results}
To evaluate the discriminateness of soft-biometric attributes for recognition, in Section \ref{sec:FeatureImportance} the attribute importance for the recognition decision is presented.
In Section \ref{sec:RecognitoinSoftBiometrics} the verification and identification performance based on the attribute information is reported.
Section \ref{sec:Support} demonstrates how well these attributes can support hard face biometrics in verification and identification tasks.
Finally, the findings are summarized in Section \ref{sec:Summary}.

%Recognition based on soft-biometric attributes --> privacy-preserving recognition approach
%also can be used as a supporting mechanism to enhance face recognition performance
%
%explain subsections

\subsubsection{Attribute Importance}
\label{sec:FeatureImportance}

\begin{figure*}
\centering
\includegraphics[width=\textwidth]{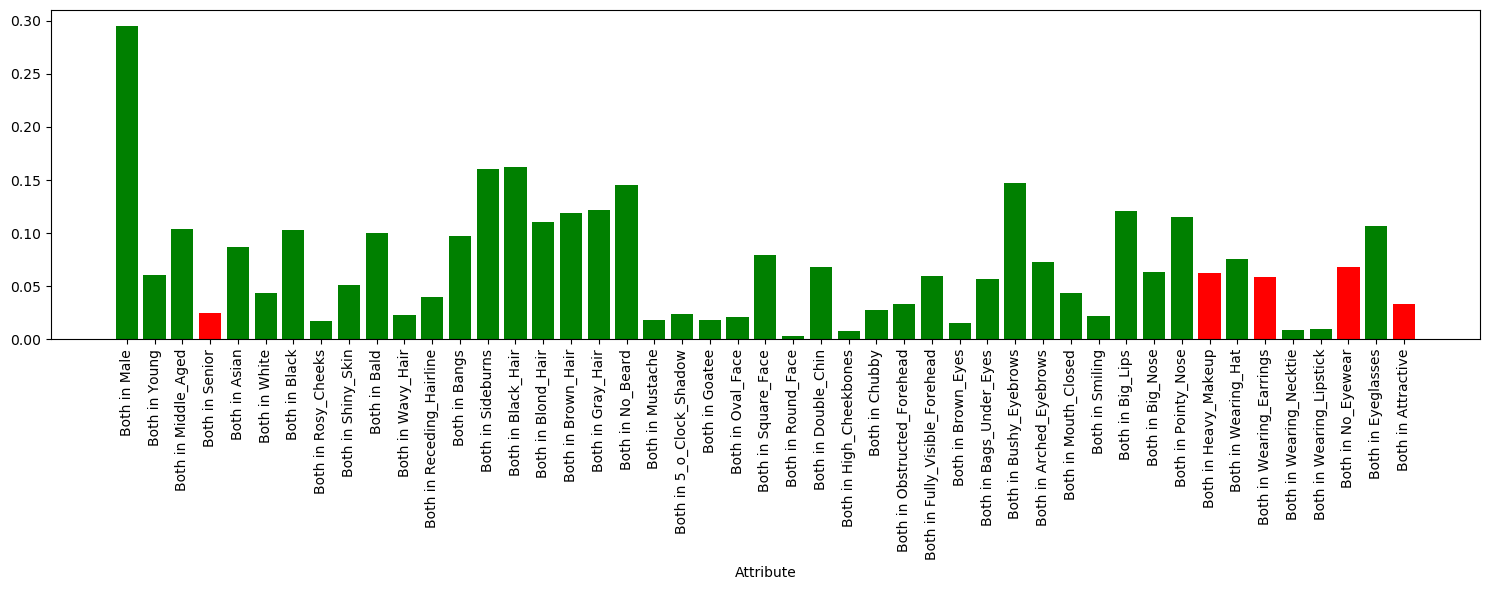}
\includegraphics[width=\textwidth]{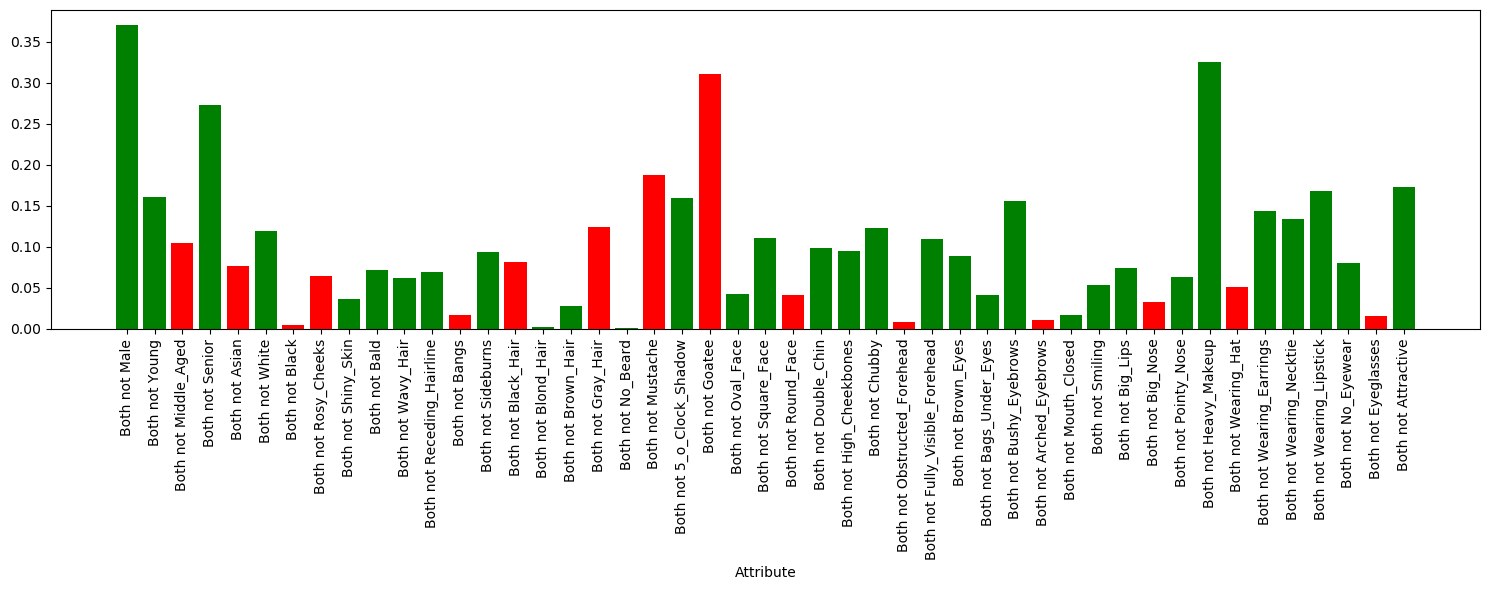}
\includegraphics[width=\textwidth]{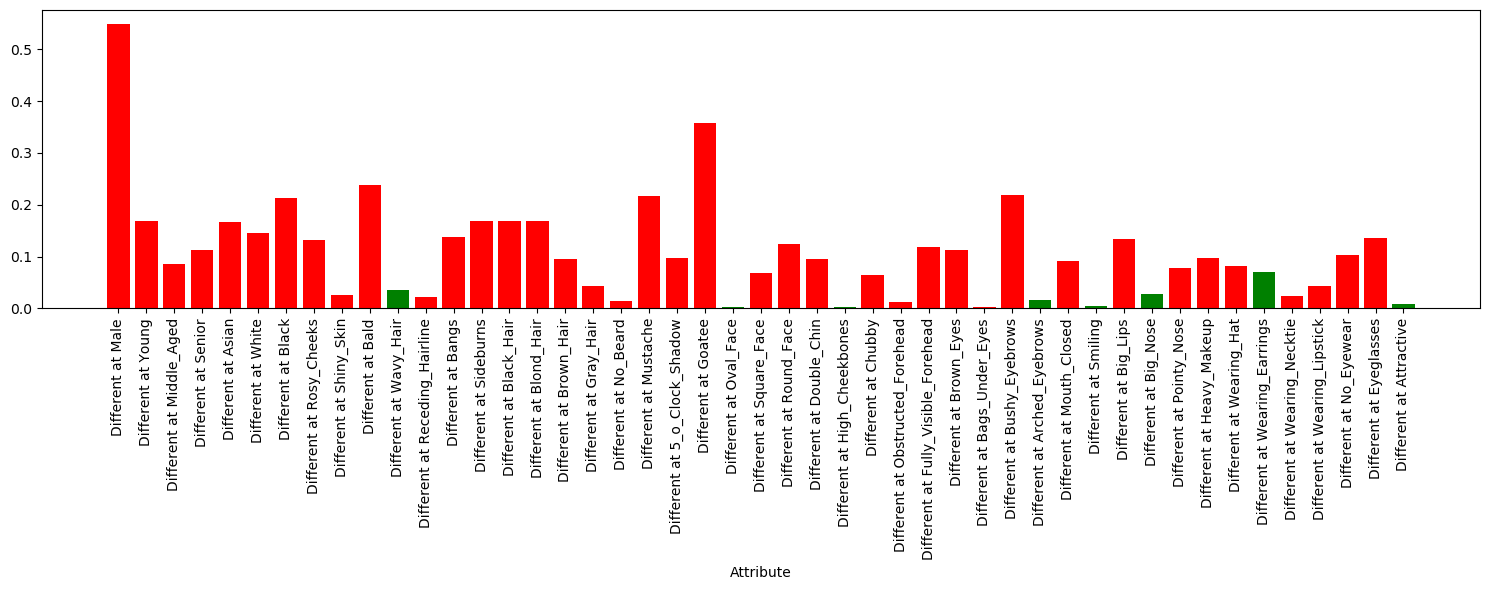}
\caption{Importance for each soft-biometric attribute derived from a logistic regression model. Green indicate the importance of genuine decisions while red indicates the importance of imposter decisions. The top figure shows the feature importance for \textit{True-True} annotations, the middle figure for \textit{False-False} annotations, and the bottom figure demonstrates the feature importance if the attribute annotation for one attribute is different (\textit{True-False}) for the probe and the reference sample.} 
\label{fig:FeatureImportance}
\end{figure*}

To get an understanding of which attributes support making accurate genuine and imposter decisions, Figure \ref{fig:FeatureImportance} shows the attribute importance derived from the logistic regression model.
A green bar refers to the contribution of an attribute for genuine decisions while a red bar indicates the contribution of an attribute for imposter decisions.
The top figure shows the feature importance for \textit{True-True} annotations, the middle figure for \textit{False-False} annotations, and the bottom figure demonstrates the feature importance if the attribute annotation for one attribute is different (\textit{True-False}) for the probe and the reference sample.

In Figure \ref{fig:FeatureImportance} it can be clearly seen that the top two figures show mostly green bars while the figure on the bottom is mostly red.
This indicates that if the probe and reference faces share the same soft-biometric attributes, it supports a genuine decision while not sharing an attribute strongly supports imposter decisions.
It turns out that the attribute "gender" has the strongest discriminative strength of all investigated attributes.
This is reasonable since (a) it is the most annotated attribute, (b) it helps to eliminate many potential candidates, and (c) it is hard to change this attribute.
But also hairstyle, haircolor, and wearing a beard have a significant impact on the recognition decision as well as more permanent factors of the face such as bushy eyebrows, big lips, and a pointy nose.
Surprisingly wearing eyeglasses strongly supports genuine decisions but not wearing eyeglasses is of no real significance.
This might be explained by the fact that only a smaller percentage of the faces in the database have glasses but if people wear glasses they usually wear them permanently.

Please note that, although the database is of significant size, these results should only be interpreted as indications since (a) the underlying annotation distributions affects the results and (b) the utilized logistic regression model might lead to oversimplified (linear) conclusions.

\subsubsection{Identity Recognition based on Soft-Biometrics only}
\label{sec:RecognitoinSoftBiometrics}
\begin{figure*}
%\captionsetup[subfloat]{farskip=5pt,captionskip=1pt}
\centering      
\subfloat[Verification \label{fig:VerificationFRPerformance}]{%
       \includegraphics[width=0.33\textwidth]{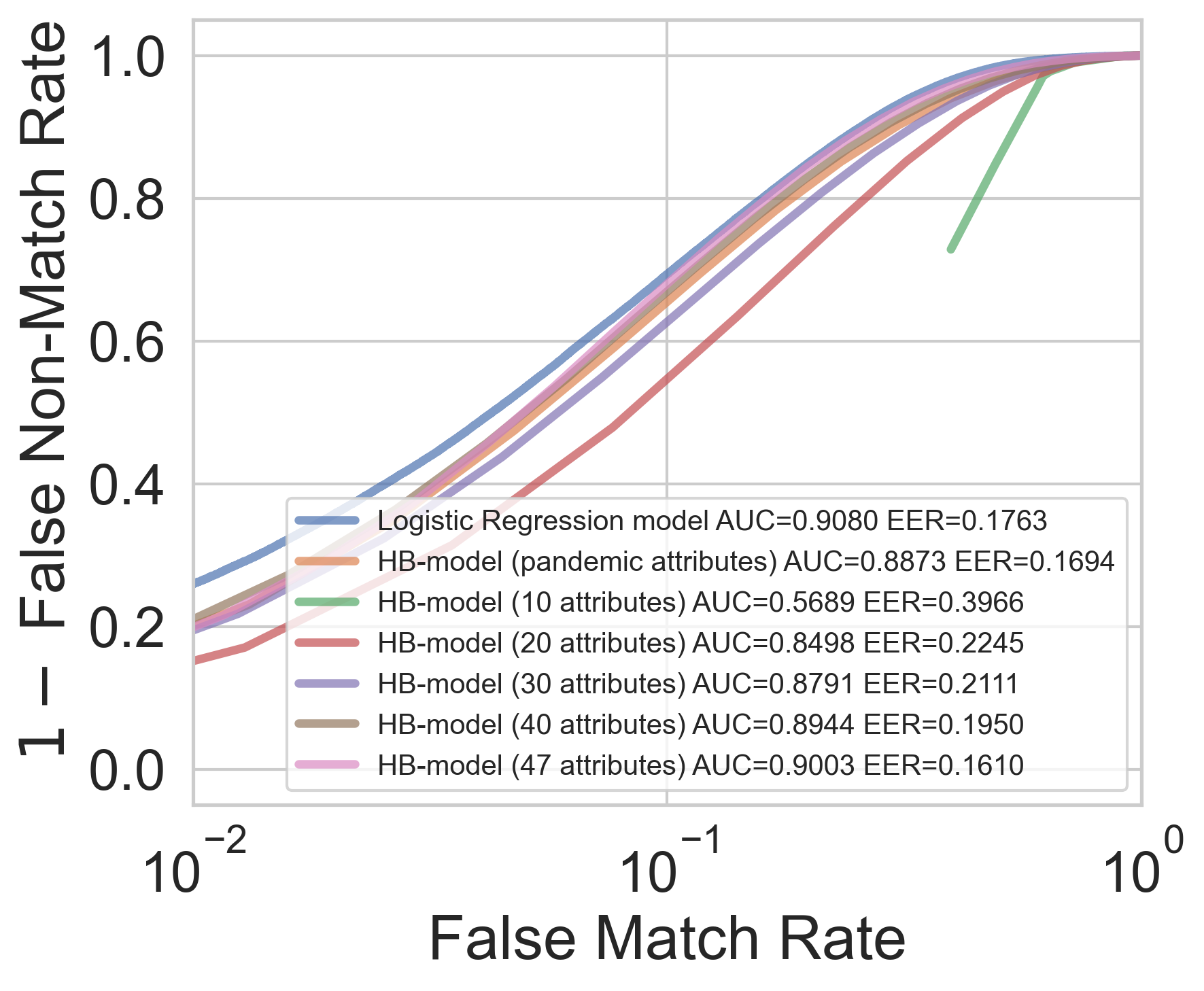}}
\subfloat[Closed-set identification \label{fig:ClosedSetFRPerformance}]{%
       \includegraphics[width=0.33\textwidth]{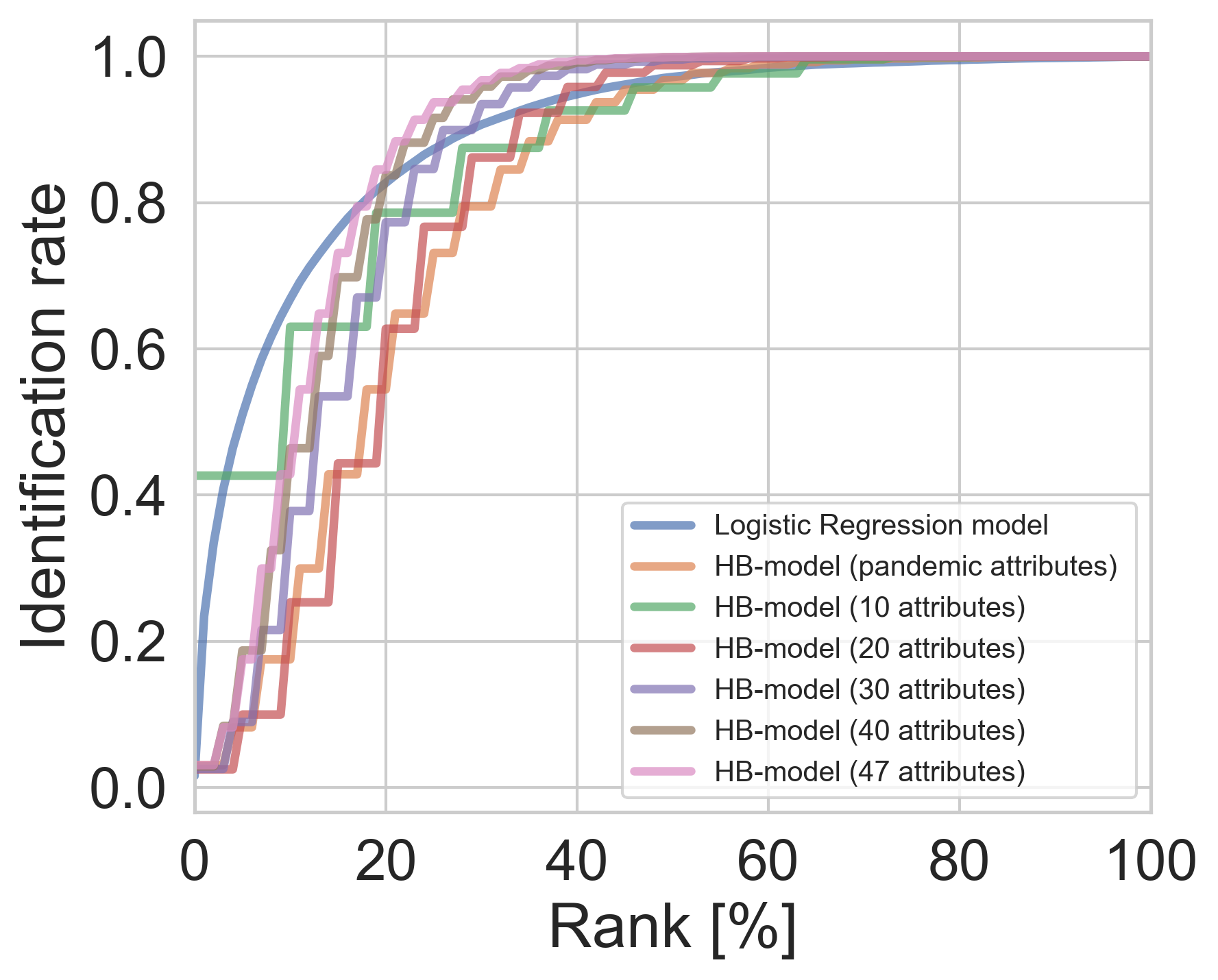}}
\subfloat[Open-set identification \label{fig:OpenSetFRPerformance}]{%
       \includegraphics[width=0.33\textwidth]{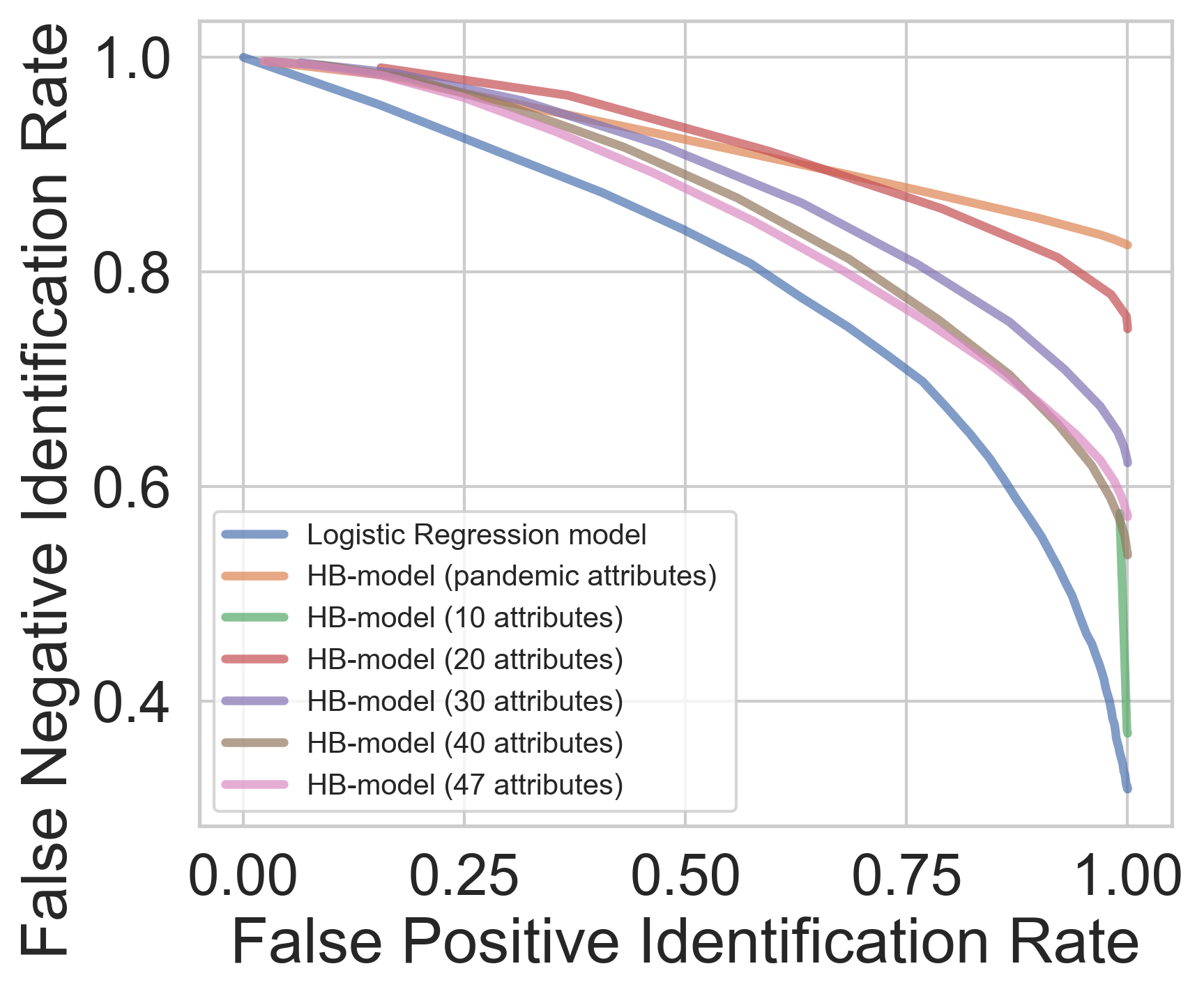}}
\caption{Analysis of the verification and identification performance based on different choices of soft-biometrics only. The verification performance is reported as an ROC curve with area under the curve (AUC) and equal error rate (EER) values. The identification performance is reported as a CMC curve for closed-set identification and as a DET curve for open-set identification. Pandemic attributes refer to attributes reliably detectable from a face in presence of a face mask.
In general, the performance increases with a higher number of considered attributes.}
\label{fig:AttributeFRPerformance}
\end{figure*}

To analyse how well soft-biometric attributes can be used for identity recognition, Figure \ref{fig:AttributeFRPerformance} shows the face recognition performance using soft-biometric annotations only.
In Figure \ref{fig:VerificationFRPerformance}, the verification performance is reported as an ROC curve including area under the curve (AUC) and equal error rate (EER) values.
Figures \ref{fig:ClosedSetFRPerformance} and \ref{fig:OpenSetFRPerformance} shows the closed- and open-set identification performance in terms of CMC and DET curves.
The plots follow the definitions of the international standards \cite{ISO_Metrik}.
The performance is reported using the logistic regression model on all attributes and using the hamming-based model on different numbers of the most relevant attributes.
Moreover, considering the current pandemic times, a special subset of soft-biometric attributes is considered that can be reliably extracted in the presence of a face mask.
This excludes attributes related to beards, mouth, nose, and face geometry \dots  as well as the attributes \textit{Wearing Lipstick}, \textit{Rosy Cheeks}, \textit{Bangs}, and \textit{Heavy Makeup}.
In total, 25 attributes are included in this subset.

The results demonstrate that it is possible to use soft-biometric attributes for both, verification and identification.
Previous works \cite{DBLP:conf/atsip/GhallebSA16, DBLP:conf/isba/AlmudhahkaNH16, DBLP:journals/pami/ReidNS14} on person identification based on soft-biometrics only reported higher performances than in our experiments.
However, these works operate on data captured in strongly controlled conditions only.
We fill this research gap by using data that were captured in strongly uncontrolled conditions and thus, posses large variations.
Most annotations used in the experiments are non-permanent and the images used for the comparisons are captured with larger time differences.
Even if such accuracy may not be enough for specific applications, we provide a starting point for further research that consider real life scenarios.
In general, a higher number of considered attributes for the hamming-based model increases the performance in all three scenarios. Moreover, the strongest performances are observed with the logistic regression based model since it is able to weight the importance of each attribute.

% new: How can soft-biometrics support hard face biometrics?
\begin{figure*}
%\captionsetup[subfloat]{farskip=5pt,captionskip=1pt}
\centering      
\subfloat[Verification \label{fig:VerificationFRSupport}]{%
       \includegraphics[width=0.33\textwidth]{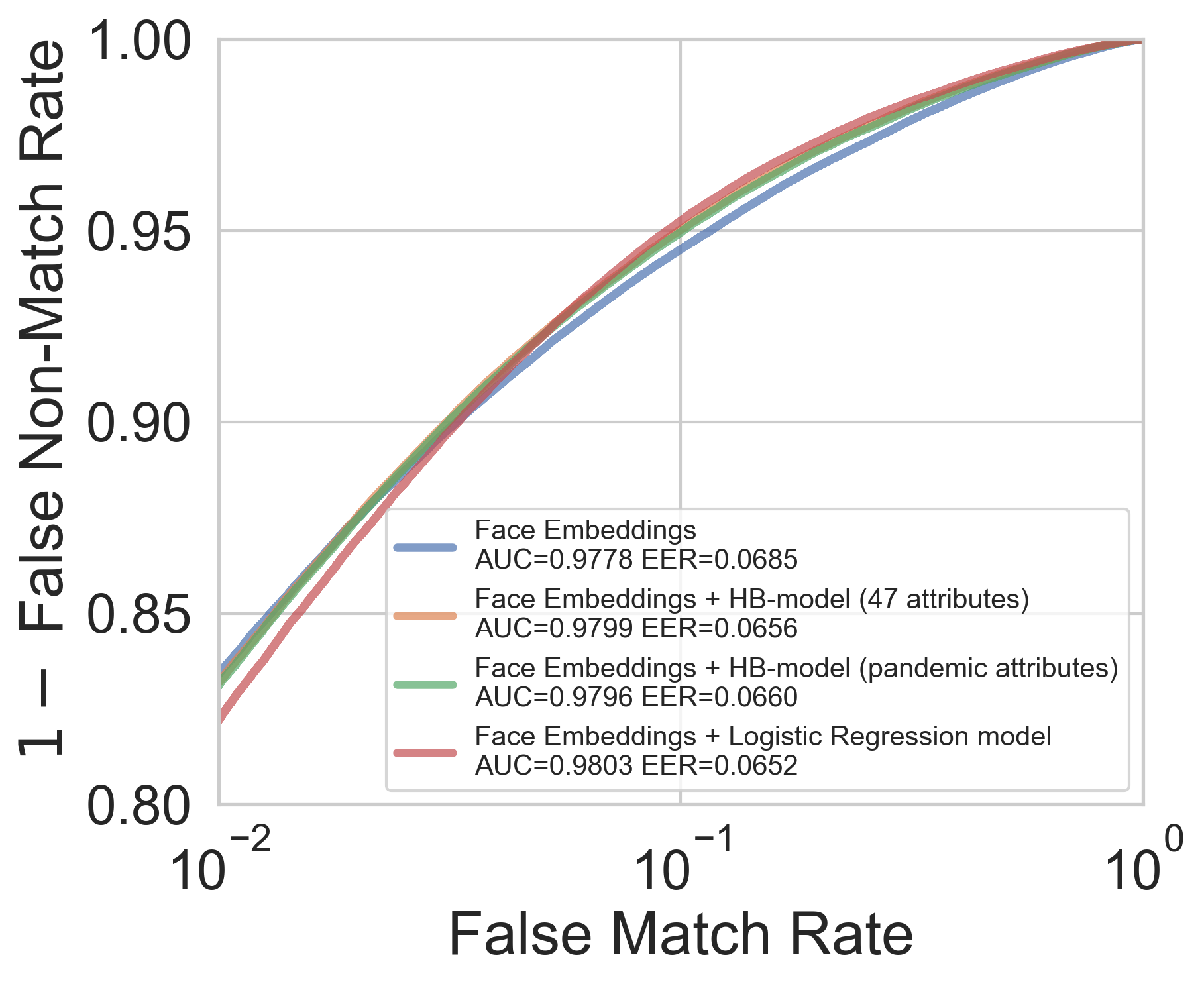}}
\subfloat[Closed-set identification \label{fig:ClosedSetFRSupport}]{%
       \includegraphics[width=0.33\textwidth]{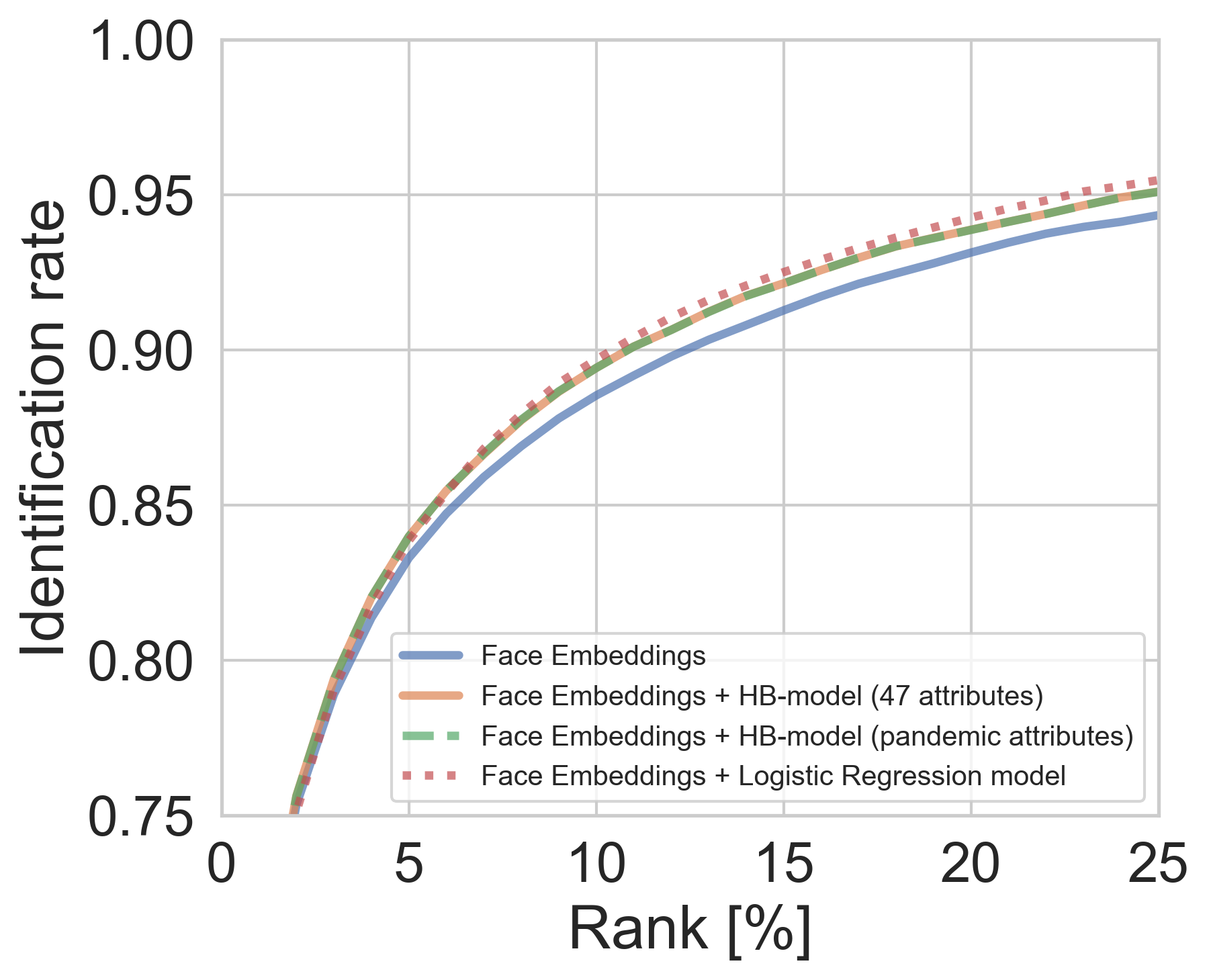}}
\subfloat[Open-set identification \label{fig:OpenSetFRSupport}]{%
       \includegraphics[width=0.33\textwidth]{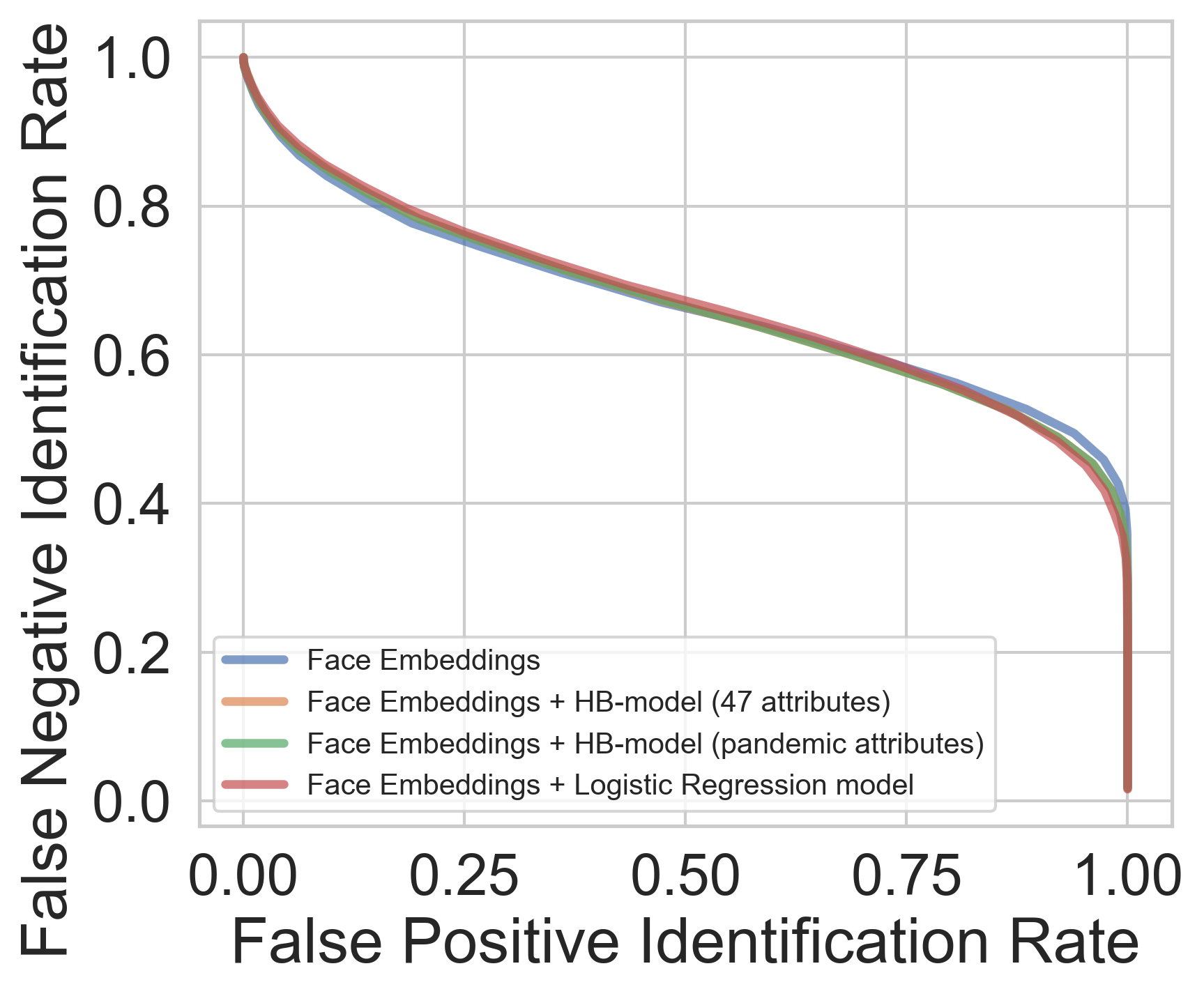}}
\caption{Analysis on how well soft-biometrics can support hard face biometrics. The verification performance is reported as an ROC curve with area under the curve (AUC) and equal error rate (EER) values. The identification performance is reported as a CMC curve for closed-set identification and as a DET curve for open-set identification. The performance using face embeddings only is reported as well as in combination with soft-biometrics. The combined approaches are achieved through a simple weighted-score fusion approach. For verification and closed-set identification, soft-biometric information enhances the performance.}
\label{fig:SupportOfSoftBiometrics}
\end{figure*}

\subsubsection{Face Biometrics Supported by Soft-Biometrics}
\label{sec:Support}
To analyse how well soft-biometrics can support hard face biometrics, Figure \ref{fig:SupportOfSoftBiometrics} shows the face verification and identification performance using face embeddings only and in combination with soft-biometrics.
The verification performance is reported as an ROC curve with area under the curve (AUC) and equal error rate (EER) values. 
The identification performance is reported as a CMC curve for closed-set identification and as a DET curve for open-set identification.
For the face embeddings, we utilized the widely-used FaceNet model\footnote{\url{https://github.com/davidsandberg/facenet}} \cite{DBLP:journals/corr/SchroffKP15}. 
The performance using face embeddings only is reported as well as in combination with soft-biometrics. 
The combined approaches are based on a simple EER-based weighted-score fusion approach as described \cite{DBLP:conf/eusipco/DamerON14}. 
The results show that for open-set identification, the score-fusion approach with soft-biometric attributes does not show significant differences compared to using the face embeddings only.
However, for the verification and closed-set identification scenarios, the additional soft-biometric information is able to enhance the recognition performance.

\subsubsection{Recognition with Soft-Biometrics - Summary}
\label{sec:Summary}
In this section, we analysed how well soft-biometrics can be used for recognition.
The annotations came from the proposed MAAD-Face annotations database.
First, we determined the attribute importance for genuine and imposter decisions.
Second, we investigated how many of these attributes are needed to achieve certain verification and identification performances given the soft-biometrics only.
Despite that the data was collected over large time-windows and the annotations are mainly of non-permanent nature, a decent verification and identification performance observed using 25 and more attribute annotations.
These results might be useful for re-identification scenarios or description-based identity search.
Lastly, we demonstrated that soft-biometrics can support hard face biometrics in verification and closed-set identification scenarios.

%recognition scenarios with small time-windows (within a day recognition) that aim at a strong user-privacy or to support general face recognition systems with a secondary soft-biometric recognition mechanism.

%database annotations not distributed as in real life scenarios
%attributes are useful for recognition in small time-windows (e.g. for recognition within a day)
%Since the data was collected over larger time windows, results only severs as indications
%and demonstrate that ...
%
%temporary attributes vs permanent attributes --> e.g. identification within small time-windows.
%
%determined the usefulness of these attributes for recognition
%
%decent verification and identification performance that can be achieved with 25 attribute and more
%--> useful for recognition scenarios with small time-windows or to strongly support face recognition systems as a secondary recognition mechanism

\section{Conclusion}
Soft-biometric attributes play a major role in the development of various face recognition topics, such as bias-mitigating, information fusion, and privacy-preserving face recognition solutions.
To support the developments in these fields, in this work, we presented four contributions.
(1) A novel annotation transfer pipeline is proposed that allows to transfer attribute annotations of high accuracy from multiple source datasets to a target dataset.
This pipeline is used to create MAAD-Face.
%Utilizing this pipeline we create the second contribution of this work, the MAAD-Face database.
(2) MAAD-Face is a novel face annotations database that provides over 3.3M faces with 123.9M annotations of 47 different attributes. 
To the best of our knowledge, MAAD-Face is the publicly available database that provides the largest number of attribute annotations.
(3) We analyse the correctness of the attribute annotations of three annotated face databases, CelebA, LFW, and MAAD-Face.
The evaluation was performed manually by three human evaluators and demonstrated that the attribute annotations of MAAD-Face are of significantly higher quality than related databases.
(4) Finally, the large number of high-quality annotations of MAAD-Face are used to study how well soft-biometrics can be used for identity recognition.
The advantage of the proposed annotation-transfer pipeline is that it allows transferring arbitrary attributes from a database to images while it ensures a high correctness of the transferred annotations.
This leads to attribute annotations of higher quality than related databases as the annotation correctness evaluation showed.
The high correctness is ensured by the use of accurate prediction reliabilities.
However, the use of this reliability might also lead to annotations correlating with specific situations in which the classifier is confident about its predictions.
This has to be addressed by future work.
We hope that this work will support the development of novel face recognition technologies.

\section*{Acknowledgment}
This work was supported by the German Federal Ministry of Education and Research (BMBF) as well as by the Hessen State Ministry for Higher Education, Research and the Arts (HMWK) within the National Research Center for Applied Cybersecurity (ATHENE), and in part by the German Federal Ministry of Education and Research (BMBF) through the Software Campus project. 
Portions of the research in this paper use the FERET database of facial images collected under the FERET program, sponsored by the DOD Counterdrug Technology Development Program Office.

\ifCLASSOPTIONcaptionsoff
  \newpage
\fi



%\bibliographystyle{IEEEtran}
% (used to reserve space for the reference number annotations box)
%\begin{thebibliography}{1}
%\end{thebibliography}

{\small
\bibliographystyle{ieee}
\bibliography{egbib}
}

\clearpage
\newpage

%\section*{Appendix}
%
%Add correlation matrix to appendix + correlation matrix of LFW and CelebA


%%\begin{IEEEbiography}{Michael Shell}
%%Biography text here.
%%\end{IEEEbiography}
%%
%%% if you will not have a photo at all:
%%\begin{IEEEbiographynophoto}{John Doe}
%%Biography text here.
%%\end{IEEEbiographynophoto}
%%
%%% insert where needed to balance the two columns on the last page with
%%% biographies
%%%\newpage
%%
%%\begin{IEEEbiographynophoto}{Jane Doe}
%%Biography text here.
%%\end{IEEEbiographynophoto}


%\vfill


\end{document}